\documentclass[20pt,twocolumn,letterpaper]{article}
\usepackage{times}
\usepackage{epsfig}
\usepackage{eso-pic}
\usepackage{enumerate}
\usepackage{graphicx}
\usepackage{amsmath}
\usepackage{amssymb}
\usepackage{subfigure}
\usepackage{algorithm}
\usepackage{algorithmic}
\usepackage{geometry} 
\usepackage{titlesec}

\titleformat{\section}
  {\normalfont\fontsize{13}{14}\bfseries}{\thesection}{1em}{}

\usepackage[pagebackref=true,breaklinks=true,letterpaper=true,colorlinks,bookmarks=false]{hyperref}
\def\abstract
   {\centerline{\large\bf Abstract}%
   \vspace*{12pt}%
   \it}

\begin{document}
%%%%%%%%% TITLE
\title{L3DOC: Lifelong 3D Object Classification}
\date{}
\author{Yuyang~Liu$^{1,2,3}$,~~~~~~Yang~Cong$^{1,2}$,~~~~~~Gan~Sun$^{1,2}$\\
$^1$State Key Laboratory of Robotics,~~Shenyang Institute of Automation,~~Chinese Academy of \\
Sciences,~~Shenyang,~~China\\
$^2$Institutes for Robotics and Intelligent Manufacturing,~~Chinese Academy of Sciences,~~\\
Shenyang,~~China\\
$^3$University of Chinese Academy of Sciences,~~China\\
{\tt\small \{liuyuyang,~congyang\}@sia.cn~~~~~~sungan1412@gmail.com}
}

\maketitle
%\thispagestyle{empty}

%%%%%%%%% ABSTRACT
\begin{abstract}
    3D object classification has been widely-applied into both academic and industrial scenarios. However, most state-of-the-art algorithms are facing with a fixed 3D object classification task set, which cannot well tackle the new coming data with incremental tasks as human ourselves. Meanwhile, the performance of most state-of-the-art lifelong learning models can be deteriorated easily on previously learned classification tasks, due to the existing of unordered, large-scale, and irregular 3D geometry data. To address this challenge, in this paper, we propose a \underline{L}ifelong \underline{3D} \underline{O}bject \underline{C}lassification (\emph{i.e.,} L3DOC) framewor, which can consecutively learn new 3D object classification tasks via imitating ``human learning''. Specifically, the core idea of our proposed L3DOC model is to factorize PointNet in a perspective of lifelong learning, while capturing and storing the shared point-knowledge in a perspective of layer-wise tensor factorization architecture. To further transfer the task-specific knowledge from previous tasks to the new coming classification task, a memory attention mechanism is proposed to connect the current task with relevant previously tasks, which can effectively prevent catastrophic forgetting via soft-transferring previous knowledge. To our best knowledge, this is the first work about using lifelong learning to handle 3D object classification task without model fine-tuning or retraining. Furthermore, our L3DOC model can also be extended to other backbone network (\emph{e.g.,} PointNet++). To the end, comparisons on several point cloud datasets validate that our L3DOC model can reduce averaged $1.68\sim3.36\times$ parameters for the overall model, without sacrificing classification accuracy of each task.
\end{abstract}

\vspace{-20pt}
%%%%%%%%% BODY TEXT
\section{Introduction}
\vspace{-5pt}
With the development of 3D perception technology, a variety of applications in 3D object recognition are developed, \emph{e.g.,} automatic driving~\cite{BehlJMARG17}, robotics~\cite{StriaH18}, bioanalysis~\cite{KalininAAFMDWHZ18}, medical diagnostic~\cite{DuanBSBDBMDOR19} and environmental perception~\cite{RoynardDG18,TangSC17}, etc. Furthermore, 3D object recognition is a classification-based recognition problem. Some 3D classification algorithms have achieved remarkably progressions, \emph{e.g.,} the well-known PointNet~\cite{QiSMG17} learns to encode each point to a global point cloud signature and PointNet++~\cite{QiYSG2017} focuses on capturing local structure in point space.
\begin{figure}[t]
\begin{center}
   \includegraphics[width=0.9\linewidth]{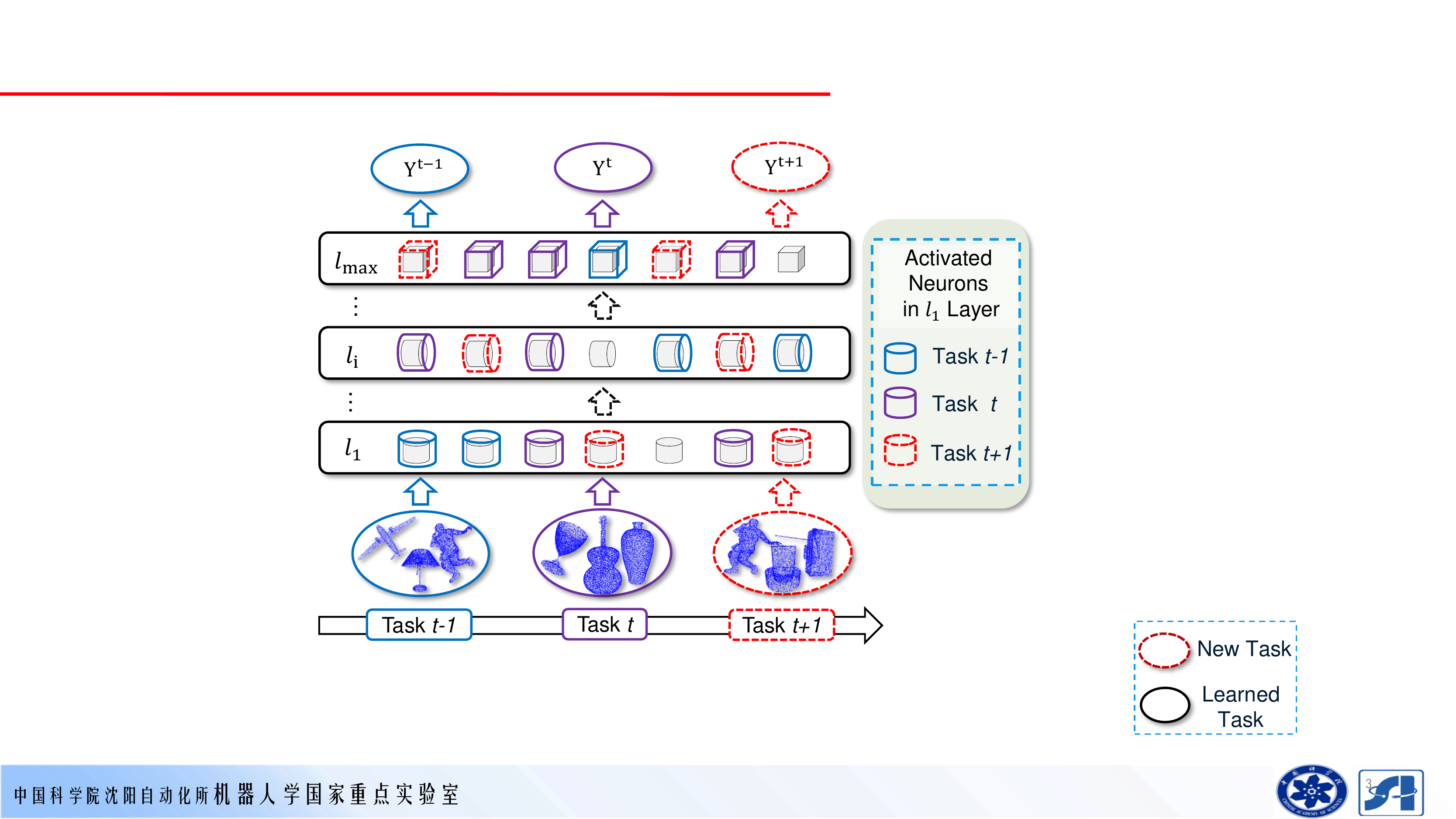}
\end{center}
\vspace{-10pt}
   \caption{Illustration of lifelong 3D object classification problem, where different shapes and colors denote different layers and object classification tasks, respectively, and dotted line denotes the new $(t+1)$-th classification task.}
   %Using the shared point-knowledge and attentional regularization mechanism, L3DOC can recognize 3D point cloud objects from unknown new task end-to-end without forgetting the performance to old ones.}
\label{fig:comparison}
\vspace{-10pt}
\end{figure}

However, as shown in Figure~\ref{fig:comparison}, most current 3D object classification models are developed for classification tasks with fixed distribution, which can result in hugely time-consuming and tedious when handling new tasks via simply retraining. Additionally, a series of new 3D classification tasks in real-world applications may have the following settings: the sequence of tasks has no definite rule, tasks are unpredictably switched, and any individual task may not be repeated frequently due to multiple environmental factors (\emph{e.g.,} illumination, occlusion and clutter). Merely fine-tuning or retraining the current model for new tasks can ignore the knowledge learned from previously learned tasks, which can cause performance degradation due to catastrophic forgetting. Therefore, this paper explores how to continually learn a series of 3D object classification tasks, while achieving less memory and higher efficiency.

A pioneering solution to these above problems is to employ lifelong machine learning methods. Generally, the fundamental goal of lifelong learning is to achieve better performance or faster convergence/training-speed for a new task, based on leveraging the knowledge of previously tasks \cite{Jaehong2018}. Although lifelong learning has been successfully applied into reinforcement learning \cite{MendezSE18, ZhanBT17} and metric learning \cite{SunCLLXY19}, most state-of-the-arts are more concerned with processing 2D data \cite{Seungwon2019, RuvoloE13}. To solve the 3D object classification problem from the perspective of lifelong learning, two main challenges need to be considered:
\vspace{-5pt}
\begin{itemize}
\setlength{\itemsep}{2pt}
\setlength{\parsep}{0pt}
\setlength{\parskip}{0pt}
  \item \textbf{Knowledge inside 3D Geometry Data:} The unordered and point-independent properties of point cloud data will lead to some existing knowledge transfer mechanisms \cite{parisi2019, Seungwon2019,kim2019deep,parisi2018, ZenkePG2017, James2017,rebuffi2017icarl} that cannot capture the shared knowledge from the geometry data according to common sense.
  \item \textbf{Catastrophic Forgetting:} Different point cloud objects can have similar approximation or geometric data among continual classification tasks, \emph{e.g.,} tables and desks in Modelnet40~\cite{WuSKYZTX15} dataset. Therefore, it is reasonable to overwrite previous task-specific knowledge to ensure the excellent adaptability of new tasks.
\vspace{-5pt}
\end{itemize}
%has similar approximation or geometric data stacking for different point cloud objects, \emph{e.g.}tables, and desks. , previously knowledge will be overwritten.

To overcome these challenges above, we develop a lifelong 3D object classification (L3DOC) framework, which intends to capture and store the shared point-knowledge from previously tasks, and adapt to new tasks without degrading performance of the previously ones. More specifically, a layer-wise tensor-channel factorization architecture is proposed to capture the unique shared point-knowledge, which can preserve the neighborhood structure of the unordered point cloud data for each task. A memory attention mechanism is employed to recognize the unknown differences between tasks, and selectively soft-transfer task-specific knowledge to learning new classification tasks. Therefore, our proposed L3DOC model can learn new point cloud object environments end-to-end with more compact parameters, while retaining the performance of the past ones. Finally, we evaluate our model by factorizing the well-known PointNet~\cite{QiSMG17} network and use it on three public point cloud benchmarks for 3D object classification, and ours have faster efficiency and excellent robustness without causing catastrophic forgetting. Meanwhile, our framework can be easily extended to other backbone networks, \emph{e.g.,} PointNet++~\cite{QiYSG2017}.

 % more compact parameters     use the shared point-knowledge without reducing recognition accuracy for new object recognition task.

The contribution of this paper is threefold:
\vspace{-5pt}
\begin{itemize}
\setlength{\itemsep}{0pt}
\setlength{\parsep}{0pt}
\setlength{\parskip}{0pt}
  \item We propose a lifelong 3D object classification (L3DOC) framework, which can transfer previous point-knowledge to learn new classification task without sacrifice classification accuracy. To our best knowledge, this is the first work about 3D object classification from the perspective of lifelong learning.
  \item We develop a general shared point-knowledge factorization architecture at the channels of convolution kernel, which can  accommodate the point-knowledge among previously learned tasks and preserve unordered neighborhood distribution in point cloud.
       %The convolution kernel tensor consists of task-specific factors and the shared point-knowledge, which are both adjustable in volume.
  \item To further reduce catastrophic forgetting the knowledge of previously tasks, we present a memory attention mechanism to selectively pick out the task-specific knowledge and soft-transfer it to learn the new tasks. Various experiment show the efficiency and effectiveness of our proposed L3DOC model.
\end{itemize}

\section{Related Work}\label{sec:related}
\vspace{-5pt}
In this section, we review related work on 3D object classification and recognition and lifelong learning, as well as works related to preventing the catastrophic forgetting problems.

%-------------------------------------------------------------------------
\subsection{3D Object Classification and Recognition}
Generally, there are three common strategies for 3D object classification and recognition: \textbf{Feature-Based Strategy}~\cite{Pauwels2013,Ulrich2012}, which usually extracts distinctive invariant (\emph{i.e.}, image scale and rotation invariant) local features from 2D color image and back project them to 3D space; \textbf{Template-Based Strategy}~\cite{tiAl-Osaimi16,BuchKK17,Cong2019,Liu2019}, which extracts template features
from the scanning model under multi-view, and searches the optimal matching by sliding windows, \emph{e.g.,} LineMod~\cite{LIHBKN12} achieve robust 3D object classification and recognition by embedding quantized image contours and standard orientations on RGBD images; \textbf{Learned-Based Strategy}~\cite{WohlhartL15,GuptaAGM15,GuptaGAM14}, various deep learning methods have shown their superior performance in 3D object classification and recognition, \emph{e.g.,} PointNet~\cite{QiSMG17} and PointNet++~\cite{QiYSG2017} directly process raw unordered point cloud data and achieve better performance by addressing the invariance to permutations and transformations of the inputs.

Although these methods above show high performance, their real-world applications inevitably rely on a large number of invariantly distributed training data~\cite{Liu2019}. Therefore, some solutions in specific applications based on multi-task learning are proposed, \emph{e.g.,} 3D Bi-Ventricular Segmentation of Cardiac Images task~\cite{DuanBSBDBMDOR19}, 3D Action Recognition~\cite{LiangFLCPZ19, OuyangXZZYLL19}, and 3D Semantic-Instance Segmentation~\cite{PhamNHRY19}. However, most multi-task learning focus on maximizing the performance of all the tasks by sharing knowledge, which consumes a lot of memory and computation~\cite{RuvoloE13}.

\subsection{Lifelong Learning}
Lifelong learning~\cite{thrun1994,RuvoloE13} is often regarded as online multi-task learning, which focuses more on efficiently learning continuous tasks based on previous knowledge while optimizing the performance across all tasks seen so far. To share fixed network parameters, they~\cite{Caruana93,rusu2016,RanjanPC19} are widely used but are more limited by known tasks. As a promotion, some methods~\cite{Gao2018,ChenZXNSLC17,LuKZCJF17} for automatically selecting parameters and network sizes have been proposed. As more complex task relationships are learned, some frameworks~\cite{JiaBTG16,LiuLFZ17,Jaehong2018} design dynamic network extensions for task-specific parts (\emph{e.g.}, filters, and neurons), which only considers the input and output spaces in general neural network architecture and ignore the critical characteristic of task-specific parts in space and channel. For the computational efficiency, \emph{i.e.}, ELLA~\cite{RuvoloE13} assumes network parameters $\theta_t$ for task $t$ is a linear combination of latent shared knowledge $L$ and task-specific sparse tensors ${s_t}$ (\emph{i.e.,} $\theta_t = L{s_t}$) \cite{RuvoloE13}. However, this linear combination cannot accommodate the knowledge embedded in deep neural network. Fortunately, the deconvolutional factorized CNN (DF-CNN)~\cite{Seungwon2019} algorithm for 2D neat data proposes a deconvolution mapping and a tensor contraction to factorize the convolution kernel:
$F^{(l)}_{t} = \mathrm{deconv}(\widehat{L}^{(l)}; V^{(l)}_{t}) \bullet U^{(l)}_{t}$,
where $\widehat{L}^{(l)}\in{\mathbb{R}^{l_h\times{l_w}\times{l_{c}}}}$ is the latent knowledge base in the $l$-th convolutional layer of each task-specific $\mathrm{CNN}_{t}$,  $V^{(l)}_{t}\in{\mathbb{R}^{v_h\times{v_w}\times{u}\times{l_{c}}}}$ is the deconvolution kernel of $\widehat{L}^{(l)}$, and $U^{(l)}_{t}\in{\mathbb{R}^{u\times{f_{in}}\times{f_{\mathrm{out}}}}}$ is the partial factor of $F^{(l)}_{t}$. When DF-CNN facing with 3D point cloud data, it inevitably causes the inefficiency since the dimension of $U^{(l)}_{t}, V^{(l)}_{t}, \widehat{L}^{(l)}$ is larger than corresponding 3D convolutional kernel. Therefore, in order to address the massive challenge for 3D geometry  data, we take the first attempt to achieve lifelong 3D object classification by storing the shared point-knowledge and selectively transferring the previous task-specific knowledge.

%-----------------------------------------------------------------------

\begin{figure*}[t]
\begin{center}
   \includegraphics[width=1\linewidth]{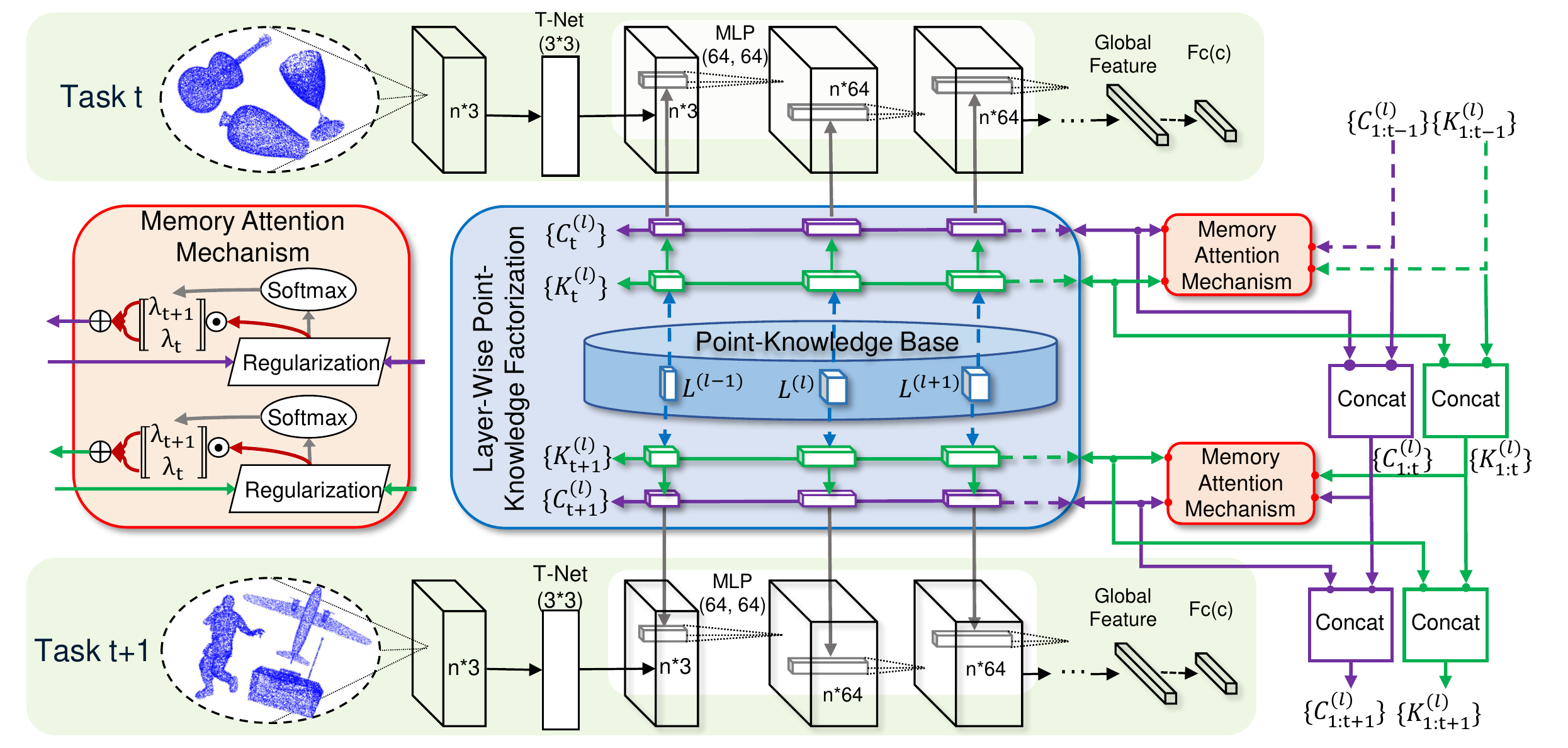}
   \vspace{-15pt}
\end{center}
   \caption{The illustration of our L3DOC framework, which shows the transfer and update of the shared point-knowledge of all the layers in $\mathcal{T}^{t},\mathcal{T}^{t+1}$. The convolution kernel $W^{(l)}$ (grey boxes) is reconstructed from the shared point-knowledge base $L^{(l)}$ (blue boxes) by learning deconvolution mapping ($K^{(l)}$ in green boxes) and tensor contraction ($C^{(l)}$ in purple boxes). Task-specific factors $K^{(l)}_{t+1}$ and $C^{(l)}_{t+1}$ in the new task is inherited from the previously tasks through a memory attention mechanism.}
\label{fig:method}
\vspace{-10pt}
\end{figure*}

\section{L3DOC: Lifelong 3D Object Classification}\label{sec:method}
\vspace{-5pt}
In this section, we comprehensively introduce our proposed lifelong 3D object classification framework. We first state the lifelong 3D object classification problem, followed by our proposed L3DOC framework.

%and estimated label $\widetilde{Y^t}\in{\mathbb{R}^{o_t\times c}}$ as outputs for all the $c$ candidate classes
%which is either directly sampled from a shape or pre-segmented from a scene point cloud.     $k_t$ scores for all the $k_t$ candidate classes
%-------------------------------------------------------------------------
\subsection{3D Lifelong Object Classification Problem}\label{sec:problem}

Given a set of $m$ 3D object classification tasks $\mathcal{T}^1,\ldots, \mathcal{T}^m$, where each individual task $\mathcal{T}^t$ consumes a set of objects $X^t\in{\mathbb{R}^{o_t}} = \{x_1^t,x_2^t,...,x_{o_t}^t\}$ and true label $Y^t\in{\mathbb{R}^{o_t\times c_t}}$ as inputs, where $x_i^t\in \mathbb{R}^{n\times d} (\forall i=1,\ldots,o_t)$ is an unordered point set, $o_t$ is the number of object for task $t$, $c_t$ is the number of class for the $t$-th task, $n$ and $d$ are the number of each point set and the dimension of each point, respectively. The corresponding output for the $t$-th task is estimated label $\widetilde{Y^t}\in{\mathbb{R}^{o_t\times c_t}}$ for all the $c_t$ candidate classes. Formally, an associated neural network PointNet~\cite{QiSMG17} (this work focuses on the classification setting, which can be extended into semantic segmentation setting) for task $t$ can be formulated as:
\begin{equation}\label{eq:pointnet}
f(X^t; \theta_t)= \rho(\mathrm{MAX}_{i = 1,..,n}\{h(x_i^t)\}),
\end{equation}
where $\theta_t$ denotes the network parameters of task $\mathcal{T}^t$, $h(\cdot)$ is approximated by a  multi-layer perceptron network, and $\rho(\cdot)$ is a continuous function.

However, continually learning 3D object classification tasks without storing the network parameters of previously learned tasks is not considered by the pioneering works. In the lifelong 3D object classification setting, the system faces with a set of 3D classification tasks $\mathcal{T}^1,\ldots, \mathcal{T}^m$ without the prior knowledge (\emph{e.g.,} task order, distribution, etc), where each $\mathcal{T}^t$ is defined in Eq.~\eqref{eq:pointnet}. The target of this system is to establish a lifelong 3D object classification framework such that: \textbf{1)} the system can ensure the classification accuracy of new tasks $\mathcal{T}^{t}$ without catastrophic forgetting the performance of old ones $\mathcal{T}^{1}...\mathcal{T}^{t-1}$; \textbf{2)} in the learning period of each new task, the convergence speed of lifelong learning system can be faster than traditional 3D object classification model, without sacrificing large memory.
%, \emph{e.g.,} PointNet~\cite{QiSMG17} and PointNet++~\cite{QiYSG2017}

\subsection{The Proposed Method}
%To transfer knowledge between different 3D recognition tasks, one naive solution is assuming parameters $\theta_t$ for task $t$ is a linear combination of latent shared knowledge $\{L^{(l)}\}^{l_{\mathrm{max}}}_{l=1}$ in $\{l\}^{l_{\mathrm{max}}}_{l=1}$ layers and task-specific sparse tensors ${S_t}$ (\emph{i.e.,} $\theta_t = \{L^{(l)}\}^{l_{\mathrm{max}}}_{l=1}{S_t}$) \cite{RuvoloE13}. However, this linear combination cannot accommodate the knowledge embedded in deep neural network. Fortunately, the deconvolutional factorized CNN (DF-CNN)~\cite{Seungwon2019} algorithm proposes a deconvolution mapping and a tensor contraction to factorize the convolution kernel:
%%\begin{equation}\label{eq:dfcnn}
%F^{(l)}_{t} = \mathrm{deconv}(\widehat{L}^{(l)}; V^{(l)}_{t}) \bullet U^{(l)}_{t},
%\end{equation}
%where $\widehat{L}^{(l)}\in{\mathbb{R}^{l_h\times{l_w}\times{l_{c}}}}$ is the latent knowledge base in the $l$-th convolutional layer of each task-specific $\mathrm{CNN}_{t}$,  $V^{(l)}_{t}\in{\mathbb{R}^{v_h\times{v_w}\times{u}\times{l_{c}}}}$ is the deconvolution kernel of $\widehat{L}^{(l)}$, and $U^{(l)}_{t}\in{\mathbb{R}^{u\times{f_{in}}\times{f_{\mathrm{out}}}}}$ is the partial factor of $F^{(l)}_{t}$.

Different from 2D neat data, each point in 3D object classification task is processed independently and identically, which makes it necessary to use the convolution kernel $W^{(l)}_{t}\in{\mathbb{R}^{1\times1\times{w_{\mathrm{in}}}\times{w_{\mathrm{out}}}}}$ in the convolution process, \emph{i.e.,} the convolution kernel in the multi-layer perceptron of PointNet~\cite{QiSMG17}. To achieve lifelong 3D object classification task, as shown in Figure~\ref{fig:method}, we first introduce how to capture the knowledge in 3D geometry data via a layer-wise point-knowledge factorization architecture, and then present a memory attention mechanism to further prevent catastrophic forgetting.

\subsubsection{Layer-Wise Point-Knowledge Factorization}\label{sec:tensor-factor}

Limited by the special distribution of neighborhood of 3D point cloud data, the size of convolution kernel is set as $ (1\times1\times w_{\mathrm{in}}\times w_{\mathrm{out}})$ in PointNet~\cite{QiSMG17}, so it is necessary to decompose the convolution kernel's input and output channels in a tensor manner. Formally, each convolution kernel tensor $\mathnormal{W}^{(l)}_{t}\in{\mathbb{R}^{1\times1\times{w_{\mathrm{in}}}\times{w_{\mathrm{out}}}}}$ in the $\mathnormal{l}$-th layer of the task $\mathcal{T}^{t}$ can be decomposed into two factors:
\begin{equation}\label{eq:getW}
\mathnormal{W}^{(l)}_{t} = \sum_{k=1}^{n} C^{(l)}_{t,(1,1,k)} D^{(l)}_{t, (k, w_{\mathrm{in}}, w_{\mathrm{out}})} = C^{(l)}_{t} D^{(l)}_{t},
\end{equation}
where $C^{(l)}_{t}\in{\mathbb{R}^{1\times1\times{n}}}$ is a task-specific factor and  $D^{(l)}_{t}\in{\mathbb{R}^{n\times{w_{\mathrm{in}}}\times{w_{\mathrm{out}}}}}$ is a task-specific intermediate factor of shared point-knowledge. Intuitively, this decomposition operator allows knowledge to be shared across different tasks, where parameter $n$ can decide how important the knowledge is to current task $\mathcal{T}^t$, and be controlled by the convolution kernel output size:
\begin{equation}\label{eq:getn}
n = w_{\mathrm{out}}/\hat{n},
\end{equation}
where $\hat{n}$ is the shrinkage scale for $n\ll({w_{\mathrm{in}}}\times{w_{\mathrm{out}}})$, and can be used as adjustments for retaining task-specific information. Then the intermediate factor $D^{(l)}_{t}$ can be obtained from the shared point-knowledge $L^{(l)}$ by deconvolution mapping:
\vspace{-5pt}
\begin{equation}
\label{eq:getP}
\vspace{-5pt}
\begin{split}
D^{(l)}_{t} = \mathcal{D}_t(L^{(l)}; K^{(l)}_{t}),
\end{split}
\end{equation}
where $L^{(l)}\in{\mathbb{R}^{n\times{w_{\mathrm{in}}}\times{l_{\mathrm{out}}}}}$, $K^{(l)}_{t}\in{\mathbb{R}^{s\times{s}\times{w_{\mathrm{out}}}\times{l_{\mathrm{out}}}}}$ is the kernel of task-specific deconvolution mapping $\mathcal{D}_t(\cdot)$, and $s$ is the spatial size of $K^{(l)}_{t}$. Similarly, the volume of $L^{(l)}$ can be set according to current convolution kernel's size:
\begin{equation}\label{eq:getlout}
l_{\mathrm{out}} = w_{\mathrm{out}}/\hat{l}_{\mathrm{out}},
\end{equation}
where $l\ll{w_{\mathrm{out}}}$. Since adjusting the size of $\hat{l}_{\mathrm{out}}$ can change the storage capacity of point-knowledge between different classification tasks, the parameters of these variables can directly affect the complexity of overall network. Notice that the task-specific tensors $\{\{K^{(l)}_{t}, C^{(l)}_{t}\}^{l_{\mathrm{max}}}_{l=1}\}^{t_{\mathrm{max}}}_{t=1}$ only represent a small fraction of all the parameters, the majority of which are contained within the shared point-knowledge base $\{L^{(l)}\}^{l_{\mathrm{max}}}_{l=1}$. Therefore, after expressing new convolution kernels $W^{(l)}_{t}$ as task-specific tensor factors $K^{(l)}_{t}, C^{(l)}_{t}$, one can learn them on $\mathcal{T}^t$ end-to-end.

\subsubsection{Factor Regularization}\label{sec:tensor-factor-loss}
Assuming we can correct various tensor parameters of one task $\mathcal{T}^t$ based on the classification loss $\mathcal{L}_{c} = \left\| \widetilde{Y}^{t} - Y^t\right\|^2_2$ in the above model, such a simple reliance on the classification loss $\mathcal{L}_{c}$ of a single task will cause the model parameters to be overcorrected and cause task-specific of shared point-knowledge base $\{L^{(l)}\}^{l_{max}}_{l=1}$ to be overwritten. Considering that $\{L^{(l)}\}^{l_{\mathrm{max}}}_{l=1}$ will be retrained when each task arrives, we only use the previously knowledge $\{L_{t-1}^{(l)}\}^{l_{\mathrm{max}}}_{l=1}$ under all the old tasks $\mathcal{T}^{1},...,\mathcal{T}^{t-1}$ to constrain $\{L^{(l)}\}^{l_{\mathrm{max}}}_{l=1}$ for new task $\mathcal{T}^t$. A simple way to prevent the learned $\{L_{t-1}^{(l)}\}^{l_{\mathrm{max}}}_{l=1}$ from being destroyed by $\mathcal{T}^t$ is to impose a $\ell_2$-norm regularizer to reduce the excessive knowledge gap between different tasks among all the layers:
\vspace{-5pt}
\begin{equation}\label{eq:lossL}
\mathcal{L}_{L} = \sum_{l=1}^{l_{\mathrm{max}}}\left\| L^{(l)}_{t-1} - \mathnormal{L^{(l)}}\right\|^2_2.
\vspace{-5pt}
\end{equation}
With the constraint above, $\{L^{(l)}\}^{l_{\mathrm{max}}}_{l=1}$ will replay previously knowledge $\{L_{t-1}^{(l)}\}^{l_{\mathrm{max}}}_{l=1}$ at all the times without forgetting the performance in $\mathcal{T}^{1},...,\mathcal{T}^{t-1}$ when the task $t$ is updated. Additionally, the performance on new task $t$ should not be lost in our L3DOC model, we thus add $\ell_2$-regularization constraints on the key task-specific tensor factors $\{\{K^{(l)}_{i}, C^{(l)}_{i}\}^{l_{\mathrm{max}}}_{l=1}\}^{t_{\mathrm{max}}}_{t=1}$ for the task $\mathcal{T}^t$:
\begin{equation}\label{eq:lossK}
\begin{aligned}
\mathcal{L}_{K_i} &=  \sum_{l=1}^{l_{\mathrm{max}}} \left\| K^{(l)}_{i} - K^{(l)}_{t}\right\|^2_2, \\
\mathcal{L}_{C_i} &= \sum_{l=1}^{l_{\mathrm{max}}}
\left\|\mathnormal{C^{(l)}_{i}} - \mathnormal{C^{(l)}_{t}}\right\|^2_2 ,
\end{aligned}
\end{equation}
where $\mathcal{L}_{K_i}, \mathcal{L}_{C_i}$ denote difference losses between the current task $\mathcal{T}^t$ and each past task $\mathcal{T}^i$, $\forall{i}=1,...,t-1$.

\renewcommand{\algorithmicrequire}{\textbf{Input:}}
\renewcommand{\algorithmicensure}{\textbf{Output:}}
\begin{algorithm}[t]
\caption{{Memory Attention Mechanism (MAM)}}
\begin{algorithmic}[1]
\REQUIRE $\{L^{(l)}_{t-1}, K^{(l)}_{1:t}, C^{(l)}_{1:t}\}^{l_{\mathrm{max}}}_{l=1}$, $\lambda\geq0$;
\ENSURE $\mathcal{L}_r$ \\
\STATE $\mathcal{L}_L \leftarrow $ Regularization($\{ L^{(l)}_{t-1}, L^{(l)}\}^{l_{\mathrm{max}}}_{l=1}$) via Eq.~\eqref{eq:lossL};
\FOR {$i=1,...,t-1$}
    \STATE $\mathcal{L}_{K_i} \leftarrow $ Regularization($ K^{(l)}_{i}, K^{(l)}_{t}$) via Eq.~\eqref{eq:lossK};
    \STATE $\mathcal{L}_{C_i} \leftarrow $ Regularization($ C^{(l)}_{i}, C^{(l)}_{t}$) via Eq.~\eqref{eq:lossK};
    \STATE  $(\lambda_{K_i}, \lambda_{C_i}) \leftarrow$AttentionScore($\mathcal{L}_{K_i},\mathcal{L}_{C_i}$) via Eq.~\eqref{eq:lambdaK};
    \STATE  $\mathcal{L}_{r}\leftarrow\lambda\mathcal{L}_L+\sum_{i=1}^{t-1}\lambda_{K_i}\mathcal{L}_{K_i}+ \sum_{i=1}^{t-1}\lambda_{C_i}\mathcal{L}_{C_i}$;
\ENDFOR
\end{algorithmic}
\end{algorithm}
\vspace{-5pt}

\subsubsection{Memory Attention Mechanism (MAM)}

In the manner described above, structural features of similar point cloud between successive tasks are stored in the shared point-knowledge base $\{L^{(l)}\}^{l_{\mathrm{max}}}_{l=1}$. To selectively transfer this knowledge and further prevent the learned $\{K^{(l)}_{i}, C^{(l)}_{i}\}^{l_{\mathrm{max}}}_{l=1}$ from losing their uniqueness, \emph{i.e.,}  catastrophic forgetting,  we add a memory attention mechanism to each $\{\mathcal{L}_{K_i}, \mathcal{L}_{C_i}\}^{t-1}_{i=1}$. The attention scores $\lambda_{K_i}, \lambda_{C_i}$ for each $\{\mathcal{L}_{K_i}, \mathcal{L}_{C_i}\}^{t-1}_{i=1}$ are set by soft-attention mechanism \cite{XuBKCCSZB15}, and can be defined as follows:
\begin{equation}\label{eq:lambdaK}
\begin{aligned}
\lambda_{K_i} &= \frac{1}{l_{\mathrm{max}}} \mathrm{\mathcal{H}}\big(
 \mathcal{L}_{K_i} | \{\mathcal{L}_{K_i}\}^{t-1}_{i=1} \big),\\
\lambda_{C_i} &= \frac{1}{l_{\mathrm{max}}} \mathrm{\mathcal{H}}
\big( \mathcal{L}_{C_i} | \{\mathcal{L}_{C_i}\}^{t-1}_{i=1} \big),
\end{aligned}
\end{equation}
where $\mathrm{\mathcal{H}(\cdot)}$ is the soft-attention mechanism obtaining the attention distribution probability information of each difference loss between $\mathcal{T}^t$ and each $\{\mathcal{T}^i\}_{i=1}^{t-1}$. The attention scores $\{\lambda_{K_i}, \lambda_{C_i}\}^{t-1}_{i=1}$ are the average attention distribution probability among all the layers, which determine how important each past tasks is to current task. Therefore, we can obtain the total loss function $\mathcal{L}_{\mathrm{total}}$ minimized in model:
\begin{equation}
\label{eq:losstotal} \vspace{-5pt}
\mathcal{L}_{\mathrm{total}} =\min \mathcal{L}_c + \lambda\mathcal{L}_L + \sum_{i=1}^{t-1} \lambda_{K_i}\mathcal{L}_{K_i} + \sum_{i=1}^{t-1} \lambda_{C_i}\mathcal{L}_{C_i}  ,
\end{equation}
where $\lambda$ controls the dependency of the shared knowledge base on past tasks. Specifically, when the difference of task-special tensors between $\mathcal{T}^i$ and $\mathcal{T}^t$ is large, the value of $\lambda_{K_i}$ and $\lambda_{C_i}$ increase, so that the attention of model optimization is paid on $\mathcal{T}^i$; when the difference of special tensors between $\mathcal{T}^i$ and $\mathcal{T}^t$ is small, the value of $\lambda_{K_i}$ and $\lambda_{C_i}$ decrease. Therefore, $\mathcal{T}^i$ cannot be considered in model optimization, and the uniqueness of $\{K^{(l)}_{i}, C^{(l)}_{i}\}^{l_{max}}_{l=1}$ within a certain range is maintained. In short, this memory attention mechanism can selectively screens out a small number of valuable task-specific tensor differences, and soft-transfer the similar knowledge among previous tasks. Finally, the goal of selectively transferring shared point-knowledge and task-specific knowledge for new task can be achieved. Furthermore, the memory attention mechanism is summarized in \textbf{Algorithm 1}, and our L3DOC framework is summarized in \textbf{Algorithm 2}.
\renewcommand{\algorithmicrequire}{\textbf{Input:}}
\renewcommand{\algorithmicensure}{\textbf{Output:}}
\vspace{-5pt}

\subsubsection{Complexity Analysis}
Meanwhile, the shared point-knowledge in $\{L^{(l)}\}^{l_{\mathrm{max}}}_{l=1}$ can be efficiently transferred to new task by retraining very few parameters while preserving the performance of previously tasks. The convolution kernels $\{W^{(l)}_t\mathrm{\in}{\mathbb{R}^{(1\times1 \times{w_{\mathrm{in}}} \times{w_{\mathrm{out}}})}}\} ^{l_{max}}_{l=1}$ of the multi-layer perceptron (\emph{i.e.}, MLP) in PointNet \cite{QiSMG17} have $N_W =  (3\times64+64^2+64^2+64\times128+128\times1024) = 159936$ parameters, and the number of additional parameters in this network are $N_{\mathrm{addition}}$. For all the trained tasks, the total number of parameters needed in STL-PointNet is:
\begin{equation}\label{eq:singletask}
\begin{aligned}
N_{\mathrm{STL-PointNet}} &= (N_{W} + N_{\mathrm{addition}}) \times t_{\mathrm{max}}\\
 &\approx N_{W}\times t_{\mathrm{max}}= 159936 \times t_{\mathrm{max}},
\end{aligned}
\end{equation}
where $N_{\mathrm{addition}} \ll N_W$. The total number of parameters in DF-CNN~\cite{Seungwon2019} based on PointNet network is:
\begin{equation}\label{eq:numberDF-CNN}
\begin{aligned}
 N_{\mathrm{DF{-}PointNet}} &= (N_{U} + N_{V})\times t_{\mathrm{max}} + N_{\widehat{L}}\\
 & = \sum_{l=1}^{l_{max}} (u \times f_{in}^{(l)} \times f_{out}^{(l)}) \times t_{\mathrm{max}} \\
 & \quad + (v_n \times v_w \times u \times l_h) \times t_{\mathrm{max}}\\
 & \quad + l_h \times l_w \times l_c\\
 & = u \times ( N_W + v_n \times v_w \times l_h)\times t_{\mathrm{max}} \\
 & \quad + l_h \times l_w \times l_c \\
 & \mathrm{>} N_W \times t_{\mathrm{max}} \approx N_{\mathrm{STL-PointNet}} ,
\end{aligned}
\end{equation}
where $f_{in}\mathrm{=}w_{in}$, $f_{out}\mathrm{=}w_{out}$, and $u \mathrm{\geq} 1$ respect to the setup in DF-CNN, which not only does not reduce the number of parameters in PointNet network, but increases the amount of additional calculations. From this perspective, the direct use of DF-CNN is not applicable in PointNet. In our L3DOC framework , the convolution kernel $\{W^{(l)}_t\}^{l_{max}}_{l=1}$ can be decomposed into smaller task-specific tensors $\{ K^{(l)}_{t}, C^{(l)}_{t}\}^{l_{\mathrm{max}}}_{l=1}$in $\mathcal{T}^t$, and shared point-knowledge $\{L^{(l)}\}^{l_{\mathrm{max}}}_{l=1}$, benefitting from the Layer-Wise Point-Knowledge Factorization. The overall parameters can be computed as:
\vspace{-5pt}
\begin{equation}
\label{eq:numberours}
\begin{split}
 N_{\mathrm{OURS}}&= (N_{C} + N_{K})\times t_{\mathrm{max}} + N_L\\
 &=\sum^{l_{\mathrm{max}}}_{l=1}\{( n^{(l)} + s^2 \times w^{(l)}_{\mathrm{out}} \times l^{(l)}_{\mathrm{out}})\times t_{\mathrm{max}} \\
 & \quad + n^{(l)} \times w^{(l)}_{\mathrm{in}} \times l^{(l)}_{\mathrm{out}}\}\\
 &=\sum^{l_{\mathrm{max}}}_{l=1}\{(\frac{1}{\hat{n}}+\frac{s^2\times w^{(l)}_{\mathrm{out}}}{\hat{l}_{\mathrm{out}}})\times w^{(l)}_{\mathrm{out}}\times t_{\mathrm{max}} \\
 & \quad +\frac{w^{(l)}_{\mathrm{in}} \times {w^{(l)}_{\mathrm{out}}}^2}{\hat{n}\times\hat{l}_{\mathrm{out}}}\}.
\end{split}
\end{equation}

\begin{algorithm}[t]
	\caption{ Lifelong 3D Object Classification Framework}
	\begin{algorithmic}[1]
		\REQUIRE $\{X^t, Y^t\}_{t=1}^{t_{\mathrm{max}}}$, $\hat{n}$, $\hat{l}_{\mathrm{out}}$, Spatial Size $s$, $\lambda\geq0$, MaxEpoch, BacthSize;
		\ENSURE $\{\widetilde{Y}^t\}_{t=1}^{t_{\mathrm{max}}}$; \\
		\STATE $\{L^{(l)}\}^{l_{\mathrm{max}}}_{l=1} \leftarrow$ Initialization($\hat{n}, \hat{l}_{\mathrm{out}}$);
		\IF {isAnotherClassificationTaskAvailable()}
		%\FOR {$t=1,...,t_{\mathrm{max}}$}
		\IF {$t=1$}
		\STATE $\{K^{(l)}_{t}, C^{(l)}_{t}\}^{l_{\mathrm{max}}}_{l=1}\leftarrow$Initialization($\hat{n},\hat{l}_{\mathrm{out}},s$);
		\ELSE
		\STATE $\{K^{(l)}_{t}, C^{(l)}_{t}\}^{l_{\mathrm{max}}}_{l=1}$ $\leftarrow$ $\{K^{(l)}_{t-1}, C^{(l)}_{t-1}\}^{l_{\mathrm{max}}}_{l=1}$; $t=t+1$;
		\ENDIF
		\STATE  BatchIndex $\leftarrow$ $Y^t/\mathrm{BatchSize}$;
		\WHILE {$i=1,\ldots,$ (MaxEpoch $\times$ BatchIndex)}
		\FOR {$l=1,\ldots,l_{\mathrm{max}}$}
		\STATE $D^{(l)}_{t} \leftarrow \mathcal{D}_t(L^{(l)}; K^{(l)}_{t}) $ via Eq.~\eqref{eq:getP};
		\STATE   $W^{(l)}_{t} \leftarrow (C^{(l)}_{t} \bullet D^{(l)}_{t}) $ via Eq.~\eqref{eq:getW};
		\ENDFOR
		\STATE $\{X^{t}_{b}, Y^{t}_{b}\} \leftarrow $ getBatchData(BatchSize, $X^t$, $\widetilde{Y}^t$);
		\STATE $\widetilde{Y}^{t}_{b} \leftarrow $BackboneNetwork($W^{(l)}_{t}, X^{t}_{b}$, $Y^{t}_{b}$);
		\STATE $\mathcal{L}_c \leftarrow$ getClassifationLoss($\widetilde{Y}^{t}_{b}, Y^{t}_{b}$);
		\IF {$t=1$}
		\STATE $\{L^{l}, K^{(l)}_{t}, C^{(l)}_{t}\}^{l_{\mathrm{max}}}_{l=1} \leftarrow$ $\mathop{\arg\min}_{L,K,C}$($\mathcal{L}_c$);
		\ELSE
		\STATE $\{L^{l}, K^{(l)}_{t}, C^{(l)}_{t}\}^{l_{\mathrm{max}}}_{l=1} \leftarrow$ $\mathop{\arg\min}_{L,K,C}$($\mathcal{L}_c$ + MAM($\{L^{(l)}_{t-1}, K^{(l)}_{1:t}, C^{(l)}_{1:t}\}^{l_{\mathrm{max}}}_{l=1}, \lambda$) );
		\ENDIF
		\ENDWHILE
		\ENDIF
		%return $\widetilde{Y}$;
	\end{algorithmic}
\end{algorithm}

From Section 3.2.1, it is known that the size of $\hat{n}, \hat{l}_{out}, s$ can affect performance of network, we thus select two better versions of parameter setting, through experiment comparison: L3DOC (\emph{Group1}) ($ \hat{n}\mathrm{=}16$, $\hat{l}_{\mathrm{out}}\mathrm{=}32$, $s\mathrm{=}2$), L3DOC (\emph{Group2}): ($ \hat{n}\mathrm{=}32$, $\hat{l}_{\mathrm{out}}\mathrm{=}32$, $s\mathrm{=}2$). Supposing $t_{\mathrm{max}}$ $\mathrm{=}$ 10,  $N_{\mathrm{OURS}}$ $\mathrm{=}$ 950664 $\mathrm{\approx}$ $ N_{\mathrm{STL-PointNet}} $ $ \mathrm{/}$ 1.68 by following L3DOC (\emph{Group1}), and $N_{\mathrm{OURS}}\mathrm{=}$475332 $\approx N_{\mathrm{STL-PointNet}} / $3.36  by following L3DOC (\emph{Group2}). Therefore, our L3DOC framework reduce 1.68$\sim$3.36$\times$ parameters less than the STL-PointNet under well performance, which is more compact and more suitable for 3D object classification tasks.

%-------------------------------------------------------
\section{Experiments}\label{sec:experiments}
\vspace{-5pt}
In this section, we carry out empirical comparisons with the state-of-the art lifelong learning models, followed by the ablation analysis about our L3DOC model.

%----------------------------------------------------
\subsection{Experimental Setup}\label{subsec:setup}
This subsection provides the detail about benchmarks, comparison models, metrics and implementation details.

\textbf{Benchmarks:}  To evaluate 3D object classification tasks, we use the point cloud benchmarks dataset, \emph{i.e.,} ModelNet~\cite{WuSKYZTX15} (ModelNet10 and ModelNet40) and 3DMNIST~\cite{3DMNIST}. Specifically, we use the Modelnet10 and Modelnet40 dataset processed in PointNet~\cite{QiSMG17} with 1024 points for each object. Corresponding to the setting of lifelong learning, we split ten 3D object classification tasks for the ModelNet10 dataset, where each task has five object categories containing training and testing data; similarly, for the ModelNet40 dataset, we randomly split it into 20 tasks, where each task 10 object categories (\emph{i.e.,} $c=c_t$ is set as 10). For 3D point cloud handwritten digits (3DMNIST), we sampled its point cloud data into (1024, 3) (\emph{i.e.,} $n=1024$ and $d=3$) by Furthest-Point-Sampling~\cite{Moenning2003} and split them into ten tasks with five object categories in each task.

\textbf{Comparison Models:}
To accurately evaluate our proposed model, we use the PointNet~\cite{QiSMG17} as backbone network and compare our model with the following models:
\textbf{1)} \emph{STL-PointNet}: Each task is trained independently based on the baseline PointNet~\cite{QiSMG17} network;
\textbf{2)} \emph{Prog-PointNet}: This model can restore the fixed PointNet~\cite{QiSMG17} network weights for each task during subsequent testing, following the progressive network's~\cite{rusu2016} setup;
\textbf{3)} \emph{DF-PointNet}: The key Deconvolutional Factorized solution in Deconvolutional Factorized CNN (DF-CNN)~\cite{Seungwon2019} is partially applied to the convolution kernel of PointNet, \emph{i.e.,} MLP. Such a setting can also compare the effectiveness of models more fairly.
\textbf{4)} \emph{Dynamically Expandable Network (DEN)}~\cite{Jaehong2018}: This model can perform the point cloud object recongnition task in an end-to-end way, which can dynamically expand the network.
\textbf{5)} \emph{L3DOC}: From Section~\ref{sec:tensor-factor}, it is known that the size of $\hat{n}, \hat{l}_{\mathrm{out}}, s$ can affect performance of network, we thus select two sets of better parameters through experiment comparison: \emph{L3DOC (Group1):} ($ \hat{n}=16$, $\hat{l}_{\mathrm{out}}=32$, $s=2$), \emph{L3DOC (Group2):} ($ \hat{n}=32$, $\hat{l}_{\mathrm{out}}=32$, $s=2$).
%Suppose $t_{\mathrm{max}}=10$, $N_{\mathrm{OURS}}= 950664\approx N_{\mathrm{STL}} / 1.68 $ by following \textbf{Group1}, and $N_{\mathrm{OURS}}=475332 \approx N_{\mathrm{STL}} / 3.36 $ by following \textbf{Group2}. Therefore, our L3DOC framework reduce 1.68$\sim$3.36$\times$ parameters less than the STL of PointNet under well performance, which is more compact and more suitable for 3D object classification tasks.

\textbf{Metrics:} The following metrics is used to demonstrate the performance on sequence tasks during the testing phase.
\begin{itemize}
    \vspace{-5pt}
    \setlength{\itemsep}{2pt}
    \setlength{\parsep}{0pt}
    \setlength{\parskip}{0pt}
  \item \emph{Peak Per-Task Accuracy (PPA)}: The Top $5\%$ test accuracy of each task in its training phase. This metric highlights the peak performance for the current task.
  \item \emph{Training Time (TT)}: The time required for completing all task training.% directly demonstrates the classification efficiency and indirectly reflects the size of the computational memory.
  \item \emph{Average Per-Task Accuracy (APA)\cite{Seungwon2019}}: The average test accuracy of all previously tasks under the current one after it has been trained, \emph{i.e.,} the overall performance of the current task model on all the tasks.
  \item \emph{Catastrophic Forgetting Ratio (CFR)}: By following \cite{Seungwon2019}, this metric denotes the ratio of the test accuracy of all seen tasks to their peak per-task accuracy.
  \item \emph{Speed of Convergence (SC)\cite{Seungwon2019}}: The convergence speed of training on each task, \emph{i.e.,} the number of required rounds when the test accuracy reaches $98\%$ of the peak accuracy of each task.
\end{itemize}

%-------------------------------------------------------------------------
\begin{figure*}[h]
	\begin{center}
		\subfigure[Average Per-task Accuracy]{
			\label{subfig:aa}
			\begin{minipage}[t]{0.32\linewidth}
				\centering
				%\caption{Average Per-task Accuracy on ModelNet10}
				\includegraphics[width=1\linewidth]{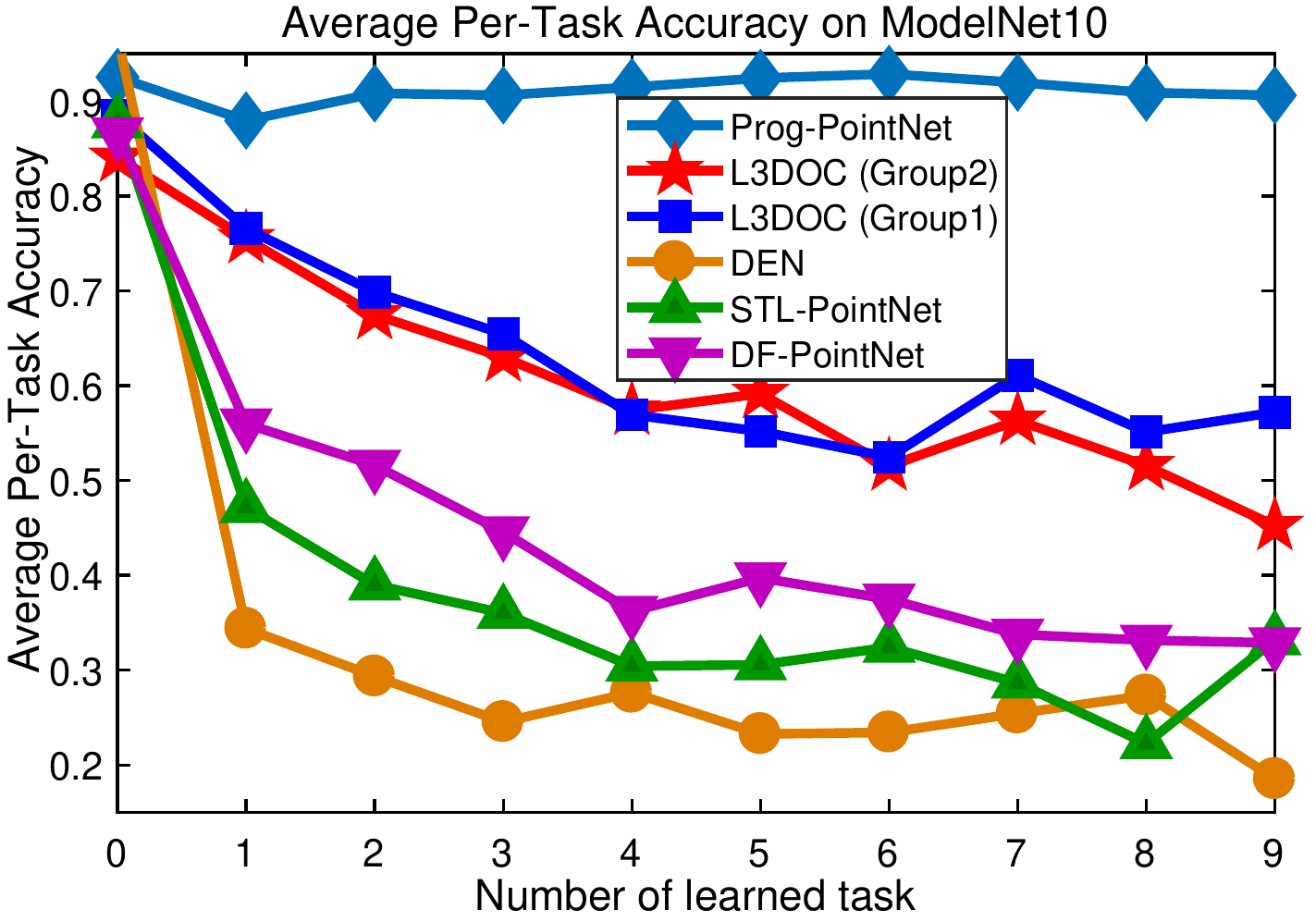}
				\vspace{4pt}
				%\subfloat[Average Per-task Accuracy on 3DMNIST]{
				\includegraphics[width=1\linewidth]{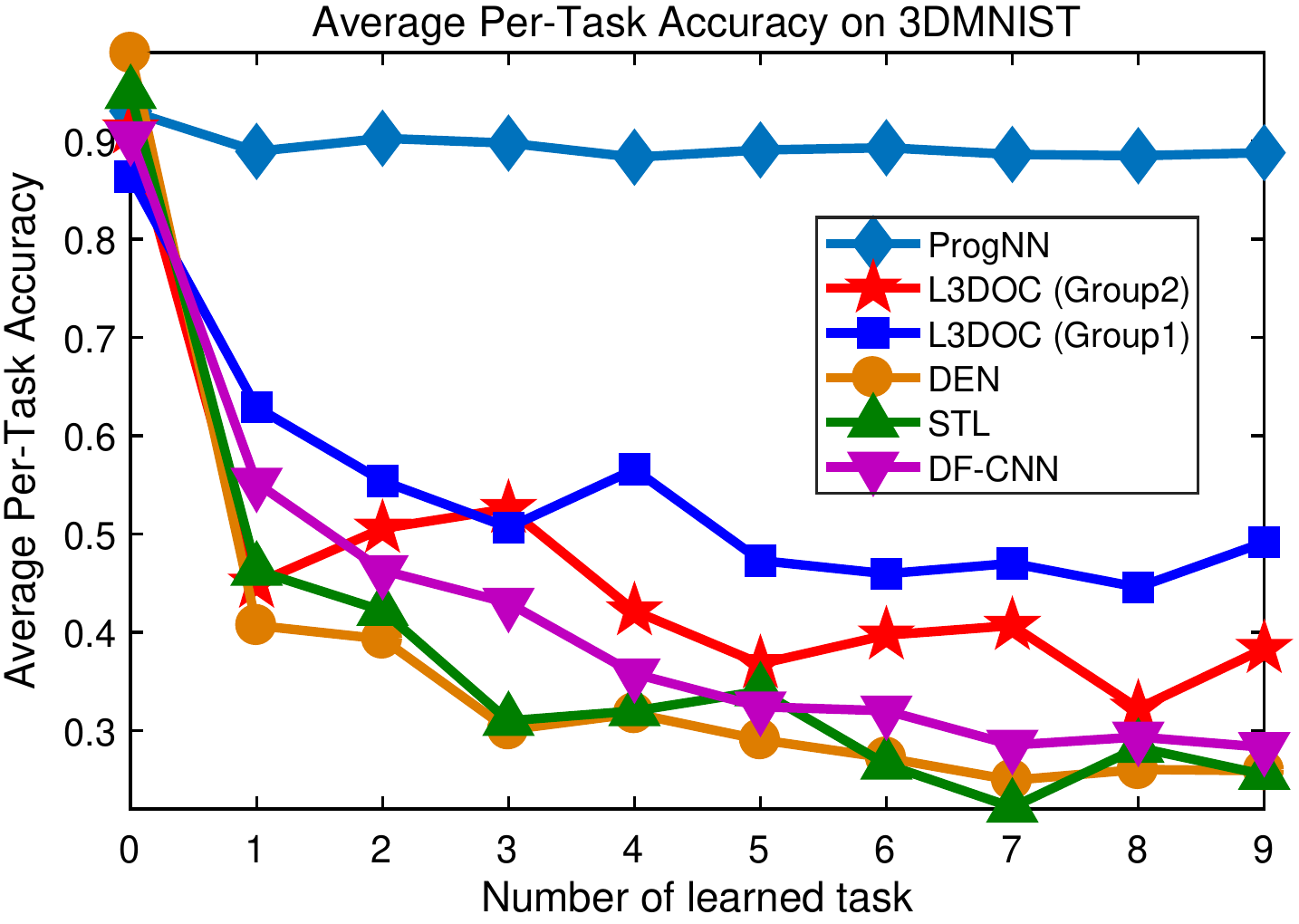}
				\vspace{2pt}
				%\caption{Average Per-task Accuracy on ModelNet40}
				\includegraphics[width=1\linewidth]{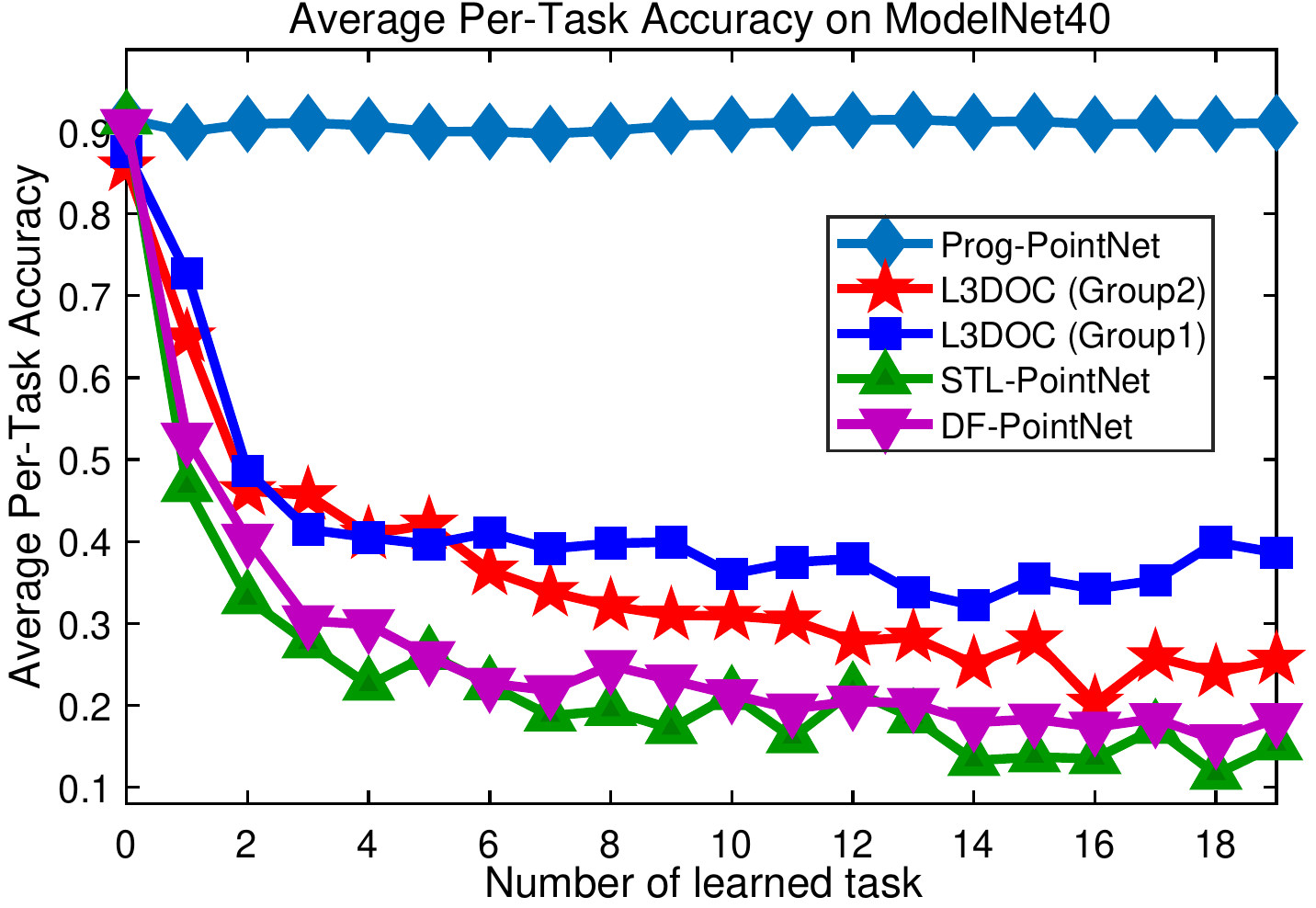}
			\end{minipage}
		}
		\subfigure[Catastrophic Forgetting Ratio]{
			\label{subfig:cfr}
			\begin{minipage}[t]{0.32\linewidth}
				\centering
				%\caption{Catastrophic Forgetting Ratio on ModelNet10}
				\includegraphics[width=1\linewidth]{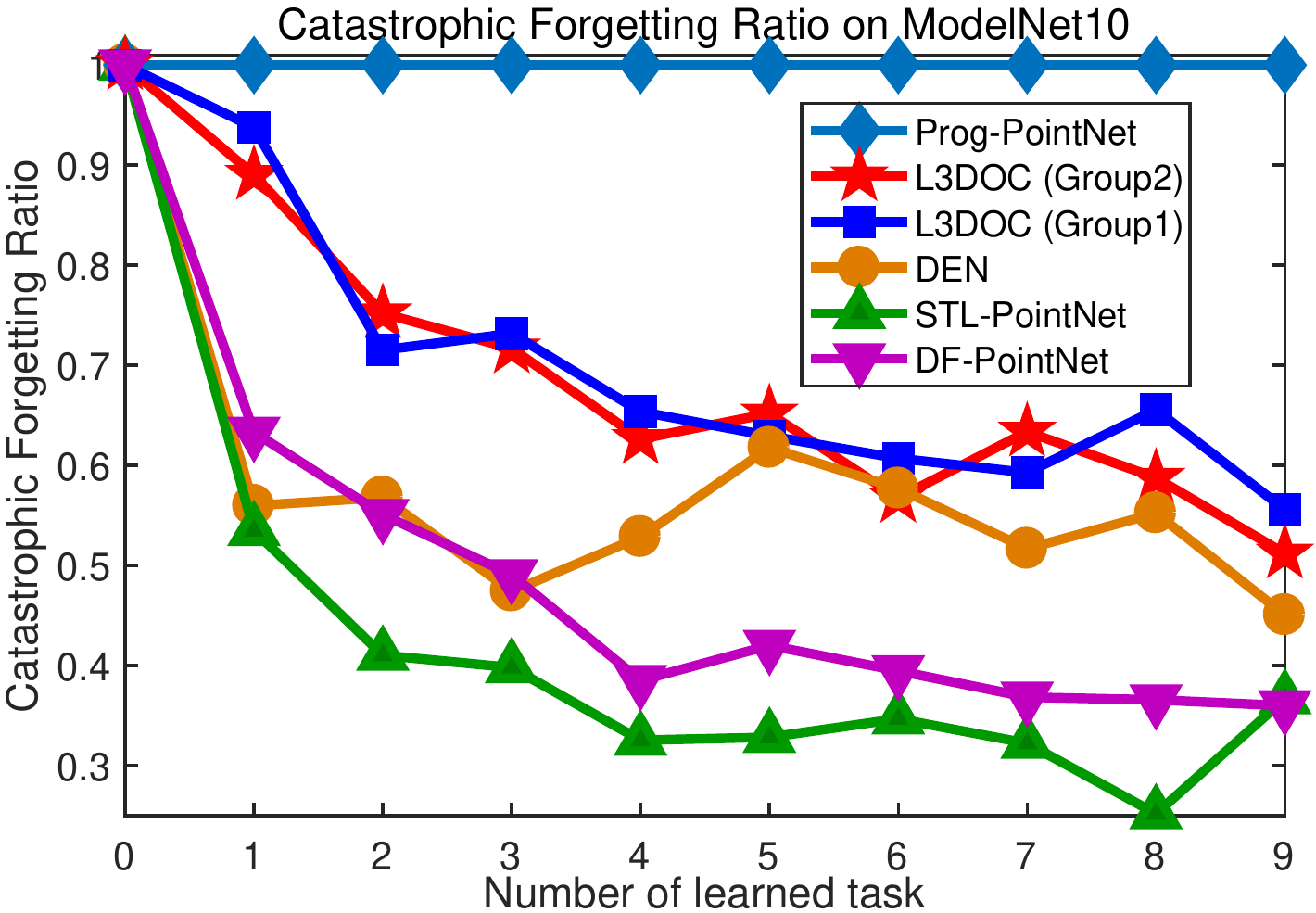}
				\vspace{4pt}
				%\caption{Catastrophic Forgetting Ratio on 3DMNIST}
				\includegraphics[width=1\linewidth]{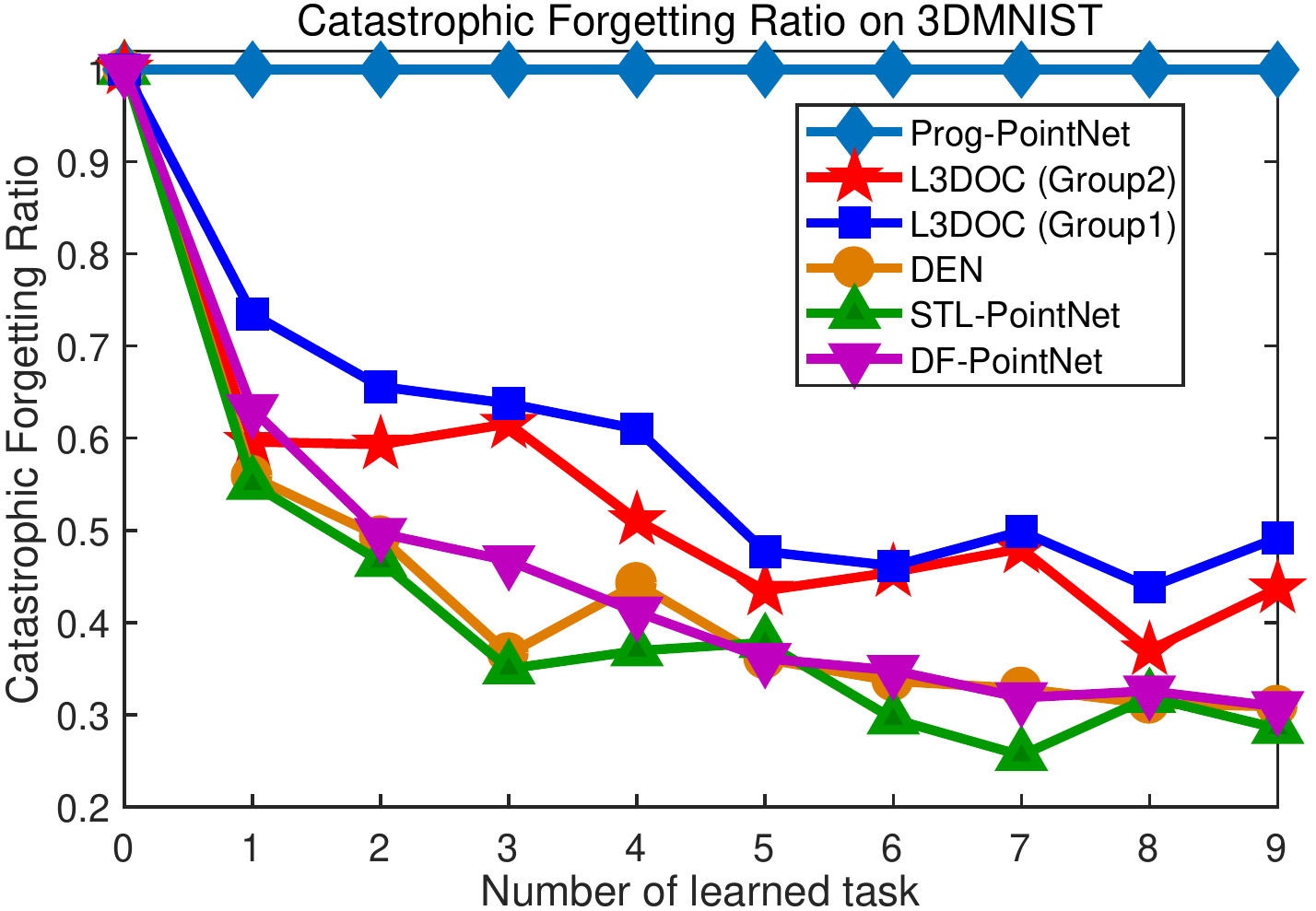}
				\vspace{3pt}
				%\caption{Catastrophic Forgetting Ratio on ModelNet40}
				\includegraphics[width=1\linewidth]{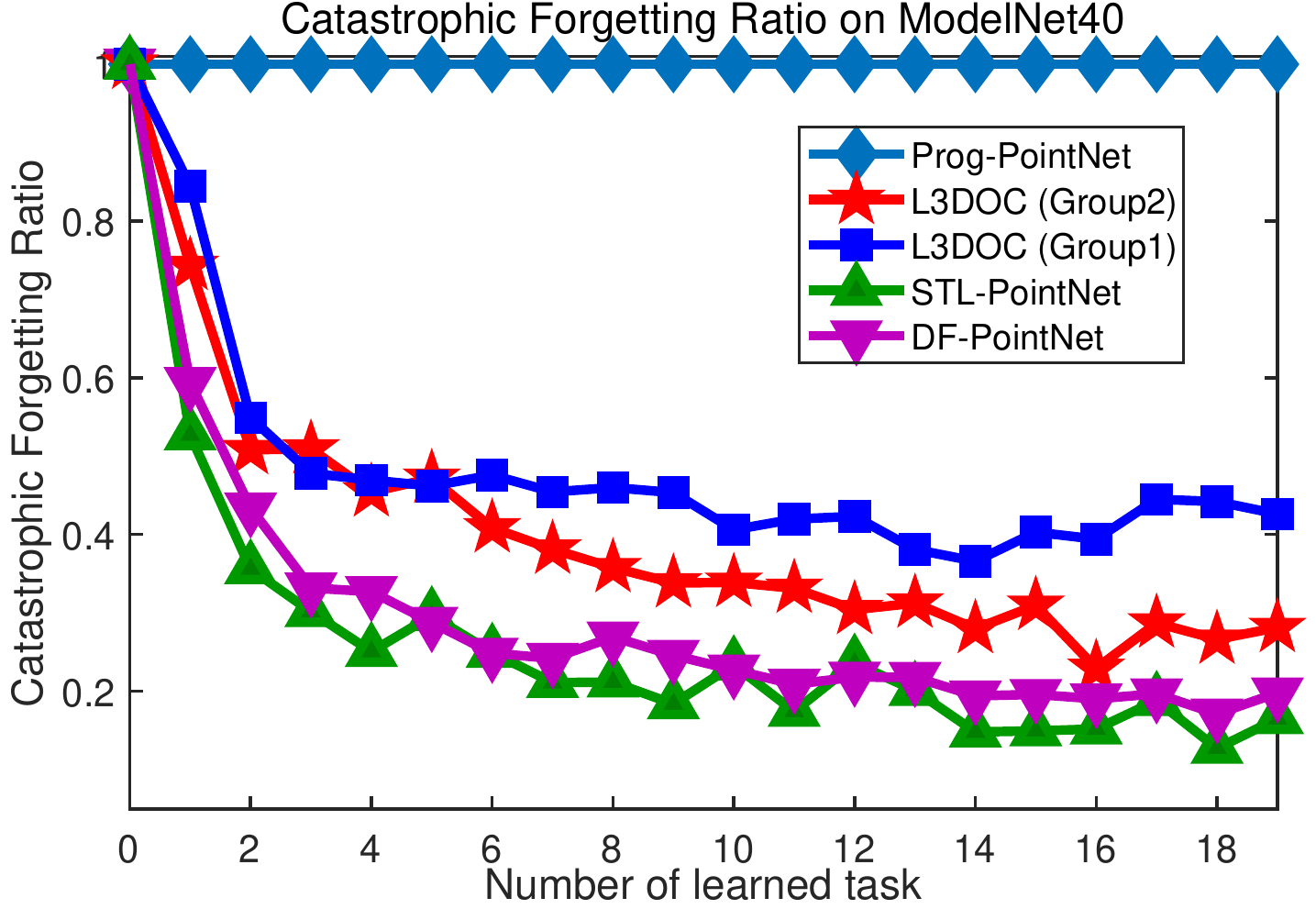}
				
			\end{minipage}
		}
		\subfigure[Speed of Convergence]{
			\label{subfig:sc}
			\begin{minipage}[t]{0.32\linewidth}
				\centering
				%\caption{Speed of Convergence on ModelNet10}
				\includegraphics[width=1\linewidth]{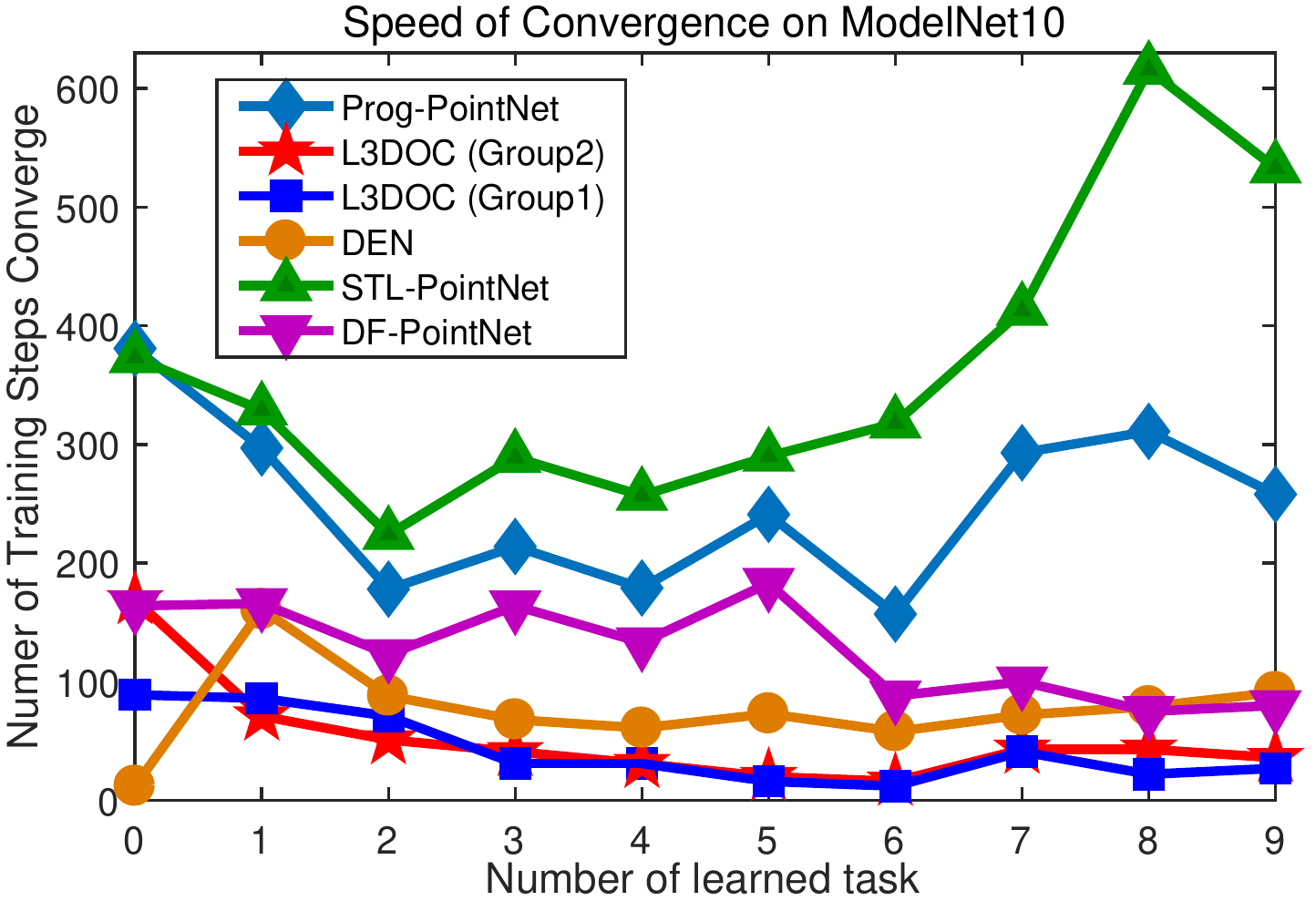}
				\vspace{2pt}
				%\caption{Speed of Convergence on 3DMNIST}
				\includegraphics[width=1\linewidth]{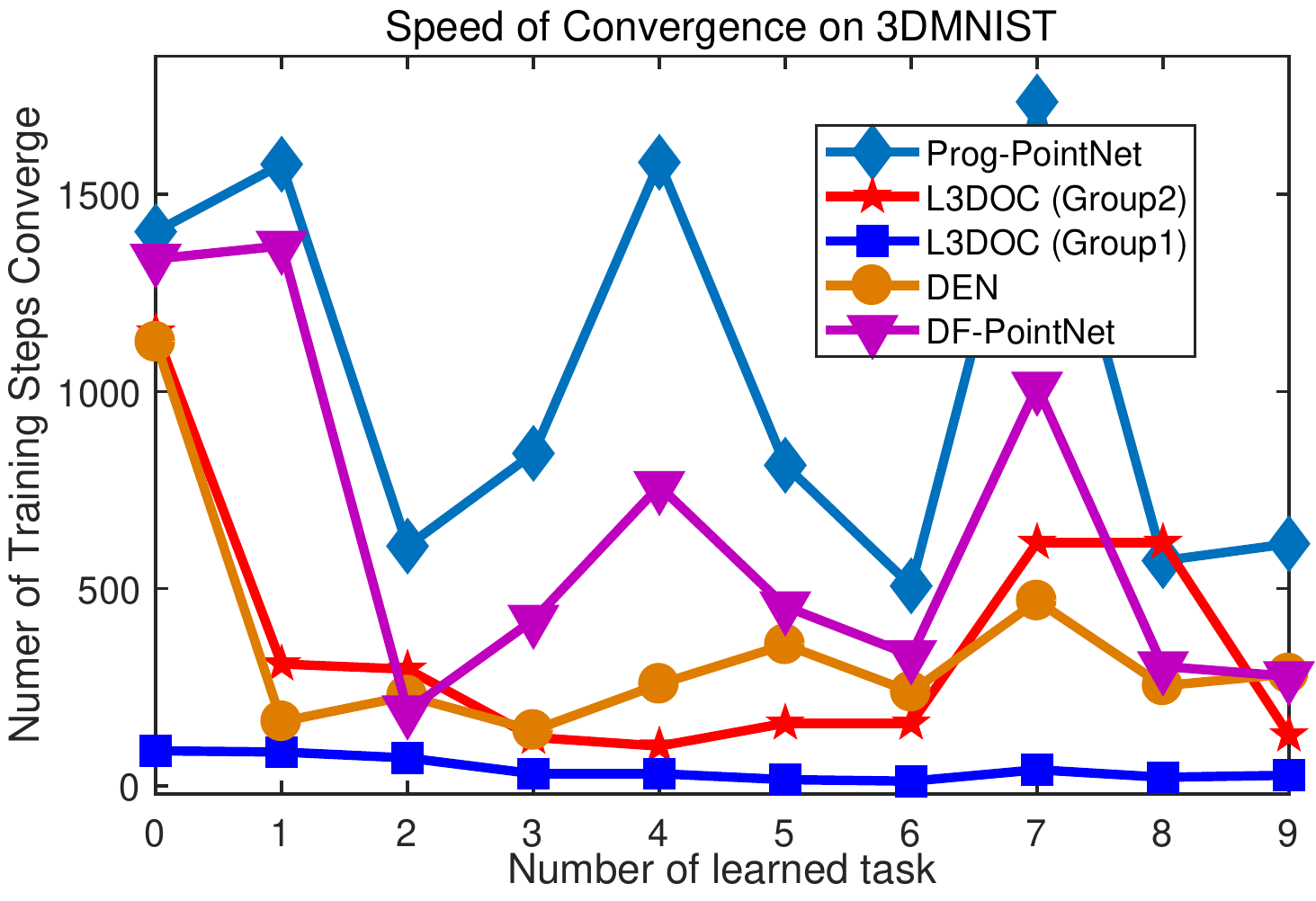}
				\vspace{2pt}
				%\caption{Speed of Convergence on ModelNet40}
				\includegraphics[width=1\linewidth]{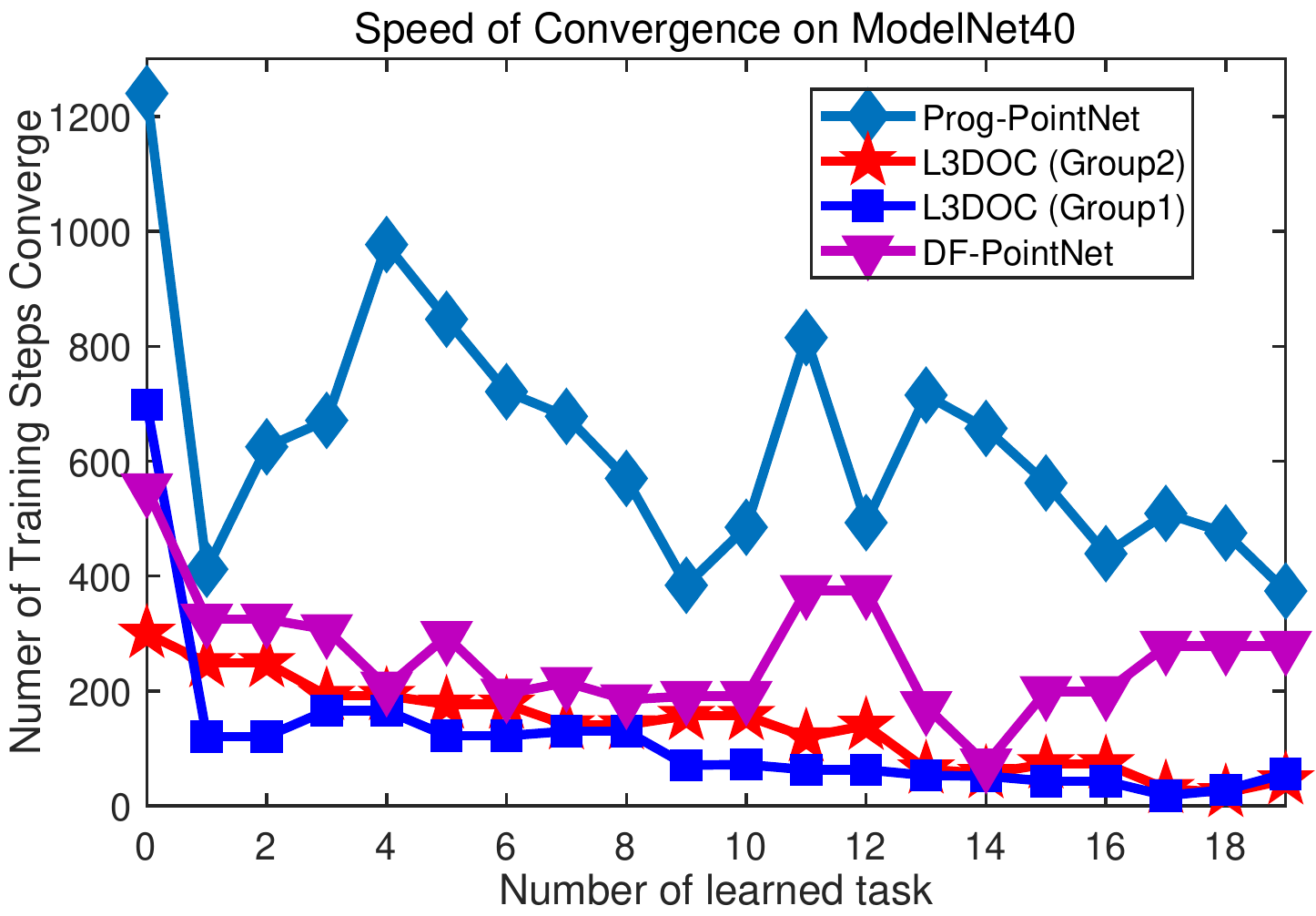}
			\end{minipage}
		}
		\vspace{-12pt}
	\end{center}
	\caption{The result of 3D object classification tasks in terms of (a) Average per-task accuracy, (b) Catatrophic forgetting ratio, and (c) Speed of convergence in three point cloud datasets (ModelNet10 (Top), 3DMNIST (Middle) and ModelNet40 (Bottom)). Lines with different colors represent different comparison models. Notice that all the metrics are decreased when increasing the number of training task.}
	\label{fig:result}
	\vspace{-5pt}
\end{figure*}

\vspace{-5pt}

\textbf{Implementation details.}  We train all models on ModelNet10 and 3DMNIST for 100 epochs, and on ModelNet40 for 160 epochs. The PointNet-based model follows its basic hyper-parameter settings. We perform the point cloud object classification task end-to-end in DEN provided by the original author. Additionally, notice that the tasks in the ModelNet40 dataset are significant differences, we do not test {DEN} model on the Modelnet40 dataset since its network size is excessively increased as a new task comes.
\vspace{-5pt}

%----------------------------------
\subsection{Experimental Results}
\vspace{-5pt}
This subsection presents experimental results in terms of efficiency and accuracy on three public point cloud benchmarks.

\vspace{-5pt}
\subsubsection{Efficiency Evaluation}
This subsection measures the efficiency \emph{v.s. PPA} of each model on all the three datasets. As the results shown in Table~\ref{tab:training}, notice that \textbf{1)} all the PointNet-based models (\emph{e.g.,} our proposed L3DOC, Prog-PointNet and DF-PointNet) can achieve excellent performance in terms of \emph{PPA}, when comparing with STL-PointNet. However, both Prog-PointNet and DF-PointNet are not faster than that of STL-PointNet, it is because they have a more high-complexity network structure. \textbf{2)} All the two versions of our model can effectively perform all the tasks while maintaining strong classification ability, which justifies the rationality of the designed layer-wise point-knowledge factorization operator.
\vspace{-10pt}
\begin{table}[h]
	\caption{Comparison the \emph{Peak Per-Task Accuracy} and \emph{Training Time} (10k second) for all the models on three benchmarks. Models with the best performance are indicated in red.}
	\vspace{-5pt}
	\begin{center}
		\footnotesize
		\setlength{\tabcolsep}{1.2mm}{
			\begin{tabular}{|l|c|c|c|c|c|c|}
				\hline
				$\#$ Benchmarks & \multicolumn{2}{|c|}{ModelNet10} & \multicolumn{2}{|c|}{ModeNet40} & \multicolumn{2}{|c|}{3DMNIST} \\
				\hline
				$\#$  Model     &  Peak-Acc & Time &  Peak-Acc & Time &  Peak-Acc & Time\\
				\hline\hline
				Prog-PointNet~\cite{rusu2016}&  100.00 & 3.21  &  100.00 & 8.09  &  100.00  & 3.66\\
				STL-PointNet~\cite{QiSMG17}  &  98.48  & 1.30  &  99.39  & 2.84  &  99.41   & 1.32\\
				DEN~\cite{Jaehong2018}	     &  53.22  & 0.42  &  - & -  & 81.11 &  0.87          \\
				DF-PointNet~\cite{Seungwon2019}  &  100.00 & 3.11  &  100.00 & 3.85  &  100.00  & 1.62\\
				L3DOC(Group1)&  99.59  & 0.89  &  100.00 & 2.88  &  98.79   & 1.29 \\
				L3DOC(Group2)&  98.19  & \textcolor[rgb]{1.00,0.00,0.00}{0.88}  &  99.80  &  \textcolor[rgb]{1.00,0.00,0.00}{2.66} &  97.39   & \textcolor[rgb]{1.00,0.00,0.00}{1.16} \\
				\hline
		\end{tabular}}
	\end{center}
	\label{tab:training}
	\vspace{-20pt}
\end{table}

%-----------------------------------------------------
\subsubsection{Accuracy Evaluation}
\vspace{-5pt}
The performance of all models is presented in Figure~\ref{fig:result}. Since knowledge in 3D geometric data is more independent than that in neat 2D data, most existing knowledge transfer algorithms do not perform as well as that on 2D data. We thus compare all the competing models with {STL-PointNet} to show the improvement of 3D classification capabilities.

For the \emph{APA} shown in Figure~\ref{subfig:aa}, we can have the following observations: \textbf{1)} except for the {Prog-PointNet}, our proposed L3DOC model can achieve the best performance on all datasets, which verifies the impact of improved generalization performance (\emph{i.e., PPA}) via layer-wise point-knowledge factorization and memory attention mechanism.
\textbf{2)} The reason why {Prog-PointNet} can obtain the best performance on the three datasets is that it retains all the parameters corresponding to each learned task, \emph{i.e.,} it can ensure the test accuracy for each task is on the corresponding training model. \textbf{3)} Comparing with {STL-PointNet}, most lifelong learning models (\emph{e.g.,} our L3DOC and {DF-PointNet}) perform better than {STL-PointNet} in terms of \emph{{PPA}}, since {STL-PointNet} cannot transfer effective knowledge from previous tasks to learn the new task. The reason why our {L3DOC} performs better than {DF-PointNet} is our {L3DOC} can capture the unique knowledge of point cloud in different objects, while retaining the global features of point cloud in the task-specific tensor parameters. \textbf{4)} For the {DEN} algorithm, it cannot achieve better performance since it pays more attention to 2D neat data while ignoring the 3D geometry data.

For the \emph{CFR} presented in Figure~\ref{subfig:cfr}, we can find that \textbf{1)} the performances of most models have similar trend as that in terms of \emph{{APA}}, \emph{i.e.,} L3DOC model can achieve the best performance in most cases. \textbf{2)} The reason why Prog-PointNet has no catastrophic forgetting is that \emph{APA} is equal to the average task test accuracy under the corresponding task training model.

\begin{figure}[t]
	\begin{center}
		\begin{minipage}[t]{0.490\linewidth}
			\centering
			%\caption{Average Per-task Accuracy on ModelNet10}
			\includegraphics[width=1\linewidth]{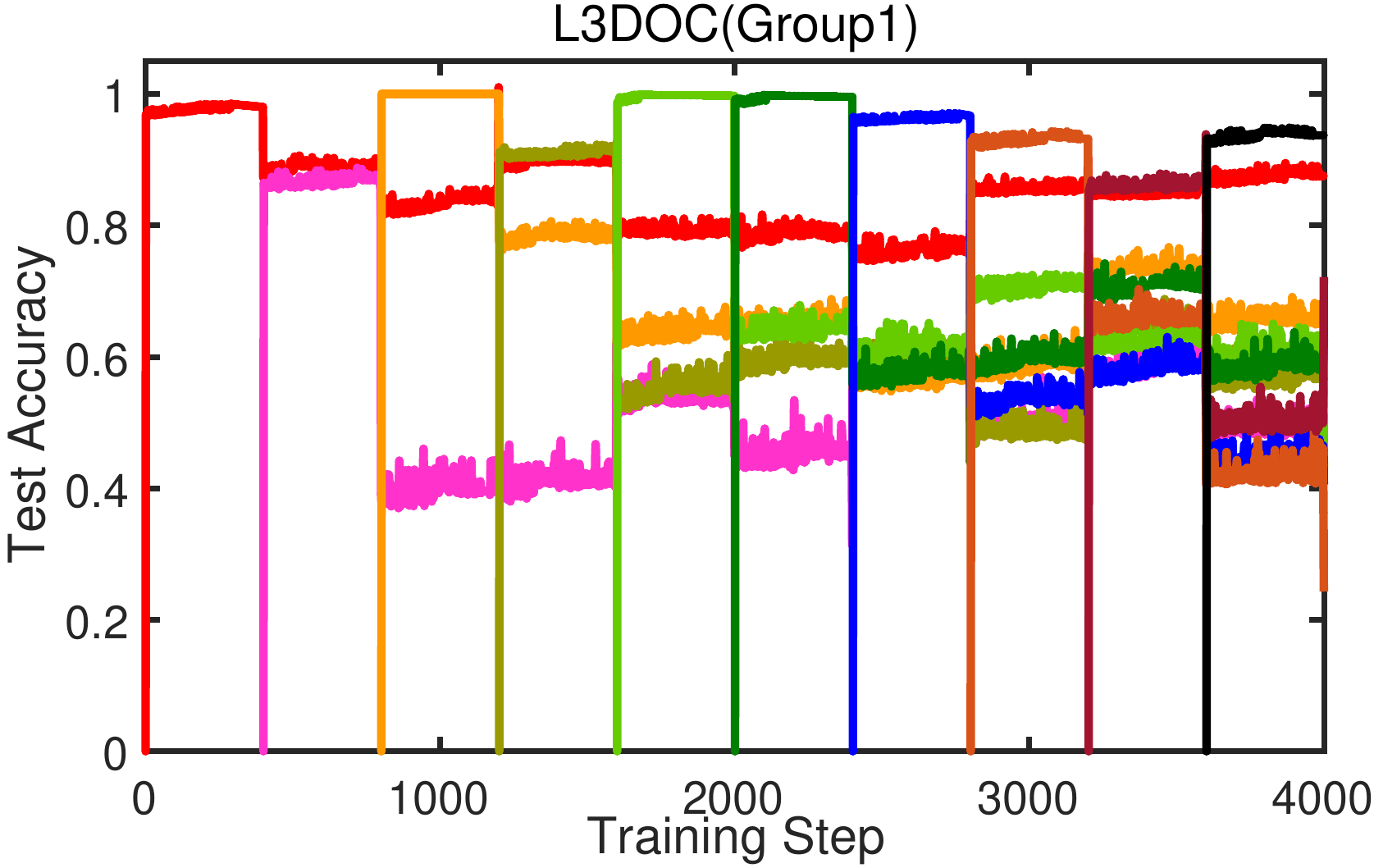}
		\end{minipage}
		\begin{minipage}[t]{0.49\linewidth}
			\centering
			%\caption{Catastrophic Forgetting Ratio on ModelNet10}
			\includegraphics[width=1\linewidth]{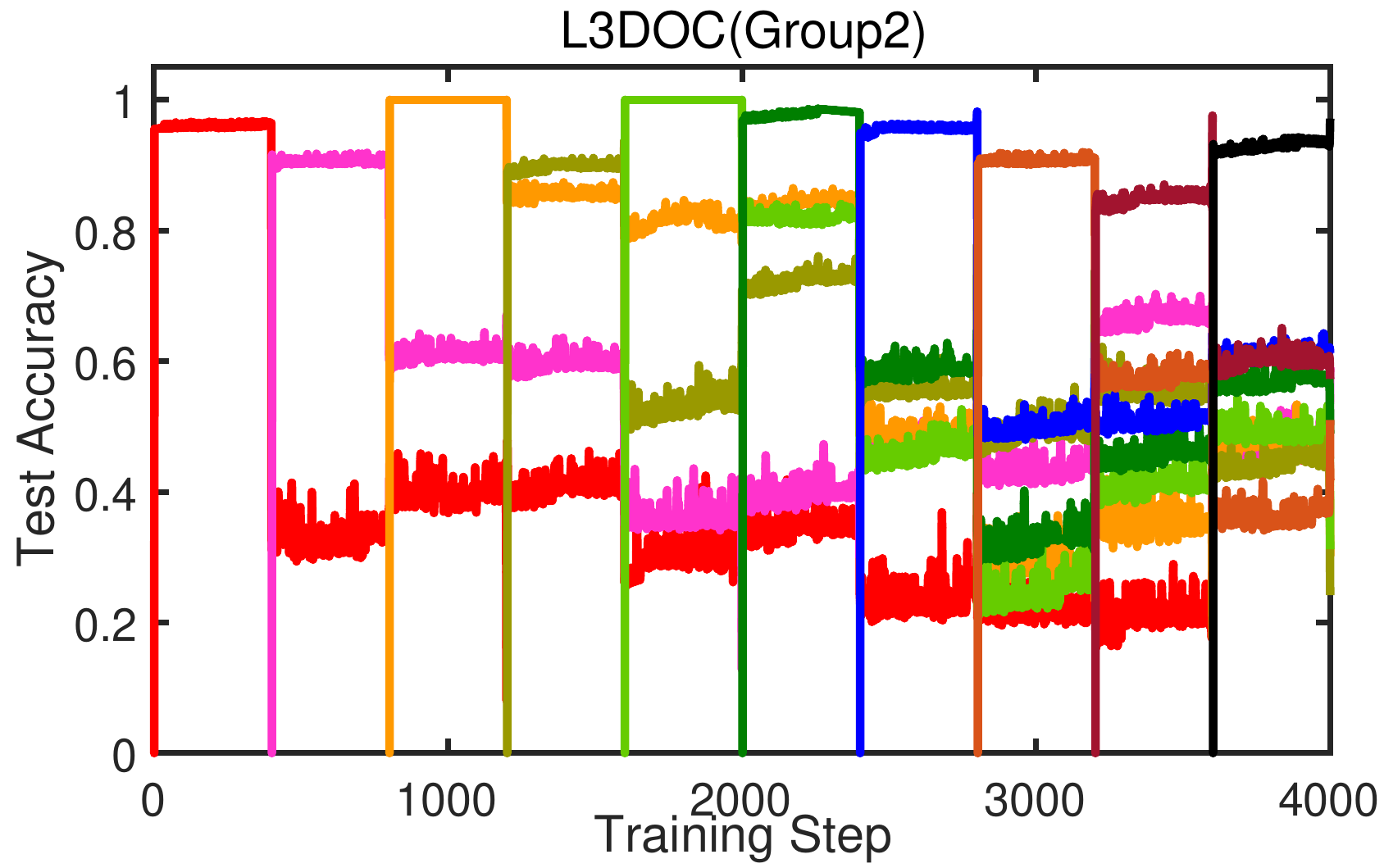}
		\end{minipage}
		\begin{minipage}[t]{0.490\linewidth}
			\centering
			%\caption{Speed of Convergence on ModelNet10}
			\includegraphics[width=1\linewidth]{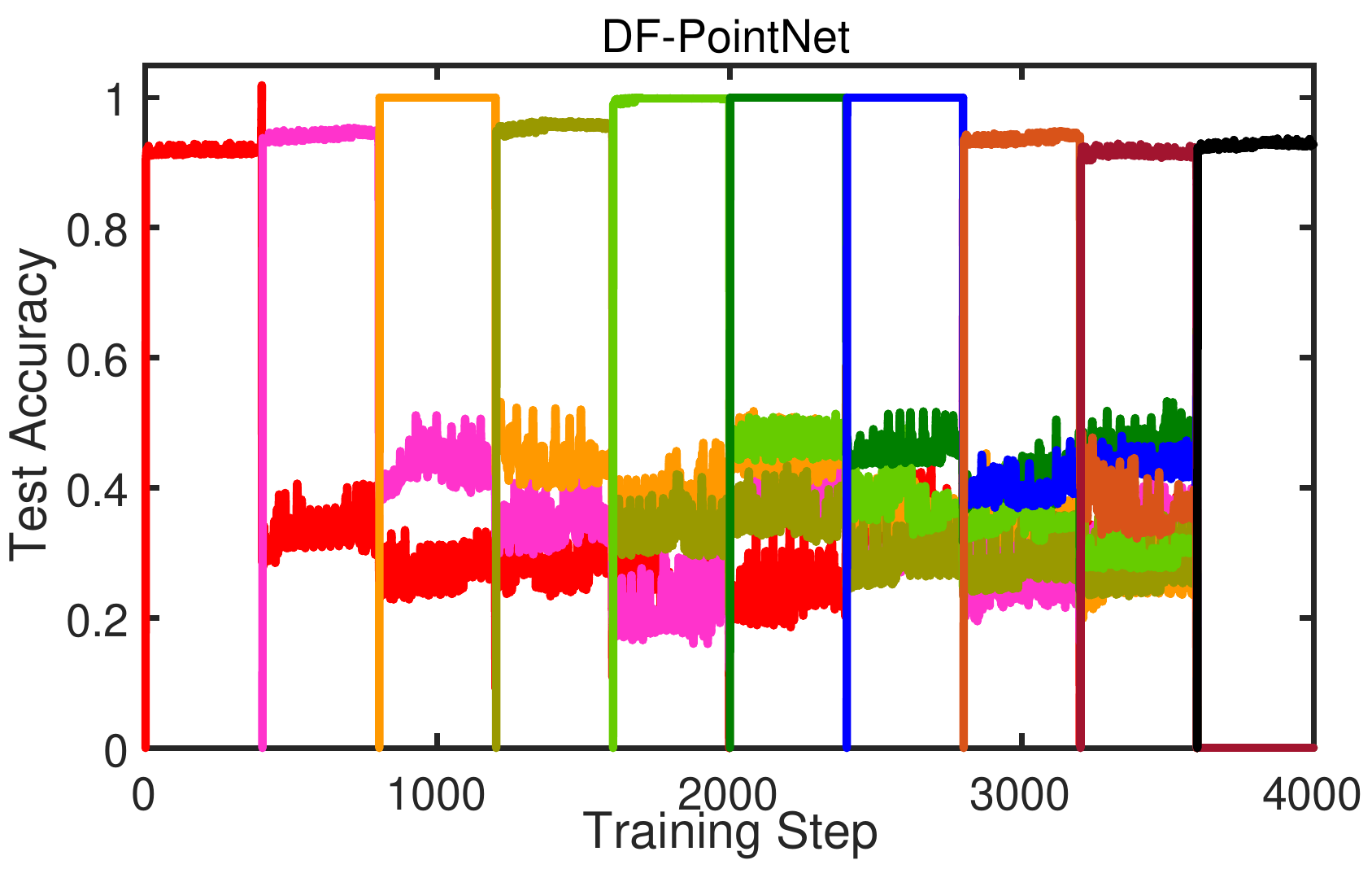}
		\end{minipage}
		\begin{minipage}[t]{0.490\linewidth}
			\centering
			%\caption{Speed of Convergence on ModelNet10}
			\includegraphics[width=1\linewidth]{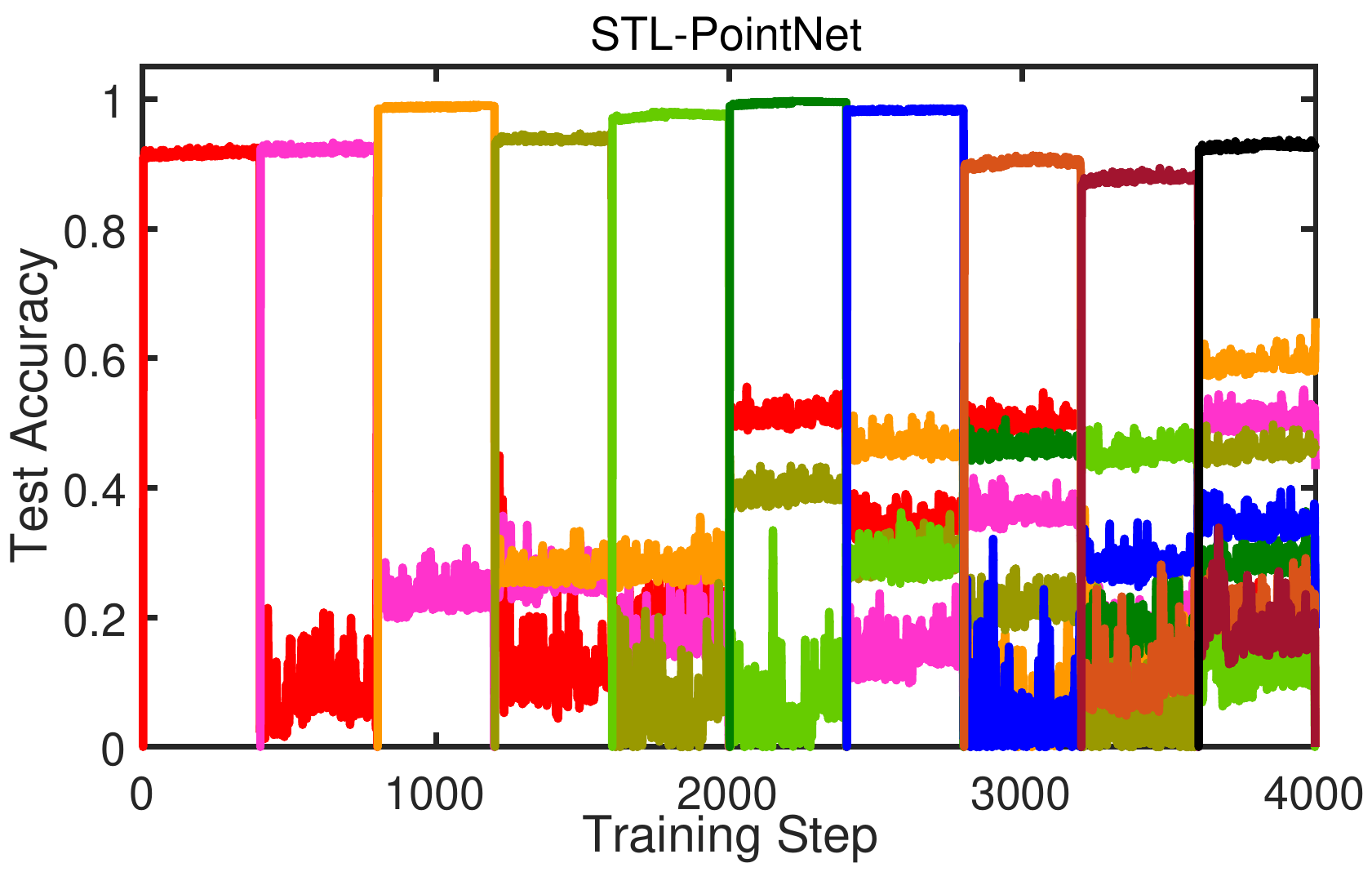}
		\end{minipage}
	\end{center}
	\vspace{-15pt}
	\caption{The test accuracy of each task during training process on ModelNet 10, where test accuracy fluctuation range is limited to $\pm10\%$ \emph{Average Per-task Accuracy}. Each color corresponds to each task that appears in a continuous way.}
	\label{fig:trainingstep}
	\vspace{-10pt}
\end{figure}

For the \emph{SC} reported in Figure~\ref{subfig:sc}, notice that \textbf{1)} our proposed L3DOC model is faster than most competing models on all the three datasets, which is in accord with complexity analysis. \textbf{2)} Although Prog-PointNet can achieve the best performance in terms of \emph{{APA}}, it needs more time to converge since it needs to retrain all the network parameters of the last task. \textbf{3)} In the 3DMNIST and ModelNet40 datasets, the \emph{Speed of Convergence} for DF-PointNet, DEN and STL-PointNet have large fluctuations. This is because that the contextual task of the fluctuating position contains objects with similar geometry but in different classes, and neither can overcome the mis-transfer of knowledge. Notice that the number of training convergence in STL-PointNet is too large, which is not indicated in Figure~\ref{subfig:sc}.

From the Figure~\ref{fig:trainingstep}, it is intuitive to see that both L3DOC(Group1) and L3DOC(Group2) can guarantee efficient test accuracy and less fluctuations on continuous tasks. Although the performance of DF-CNN is significantly improved compared with STL-PointNet (\emph{i.e.}, catastrophic forgetting), it is almost the same on all the tasks.

\vspace{-5pt}
\subsection{Ablation Studies}
This subsection studies how the scale parameters and memory attention mechanism affect the performance of our L3DOC models, and the experiment on PointNet++.

\subsubsection{Effect of Scale Parameters}
For the effect of scale parameters, as shown in Figure~\ref{subfig:aa}, Figure~\ref{subfig:cfr} and Figure~\ref{subfig:sc}, we can notice that our proposed L3DOC(Group1) model with more parameters performs well than L3DOC(Group2) model in terms of the most metrics, but consumes relatively high training time. Meanwhile, this differences in effectiveness and efficiency are more significant on the large-scale ModelNet40 dataset. It is because our L3DOC(Group1) model with more scale parameters can be flexibly adjusted to adapt into different scale tasks, while improving the scalability of the model.

\begin{figure}[h]
	\begin{center}
		%\caption{Average Per-task Accuracy on ModelNet10}
		\includegraphics[width=1\linewidth]{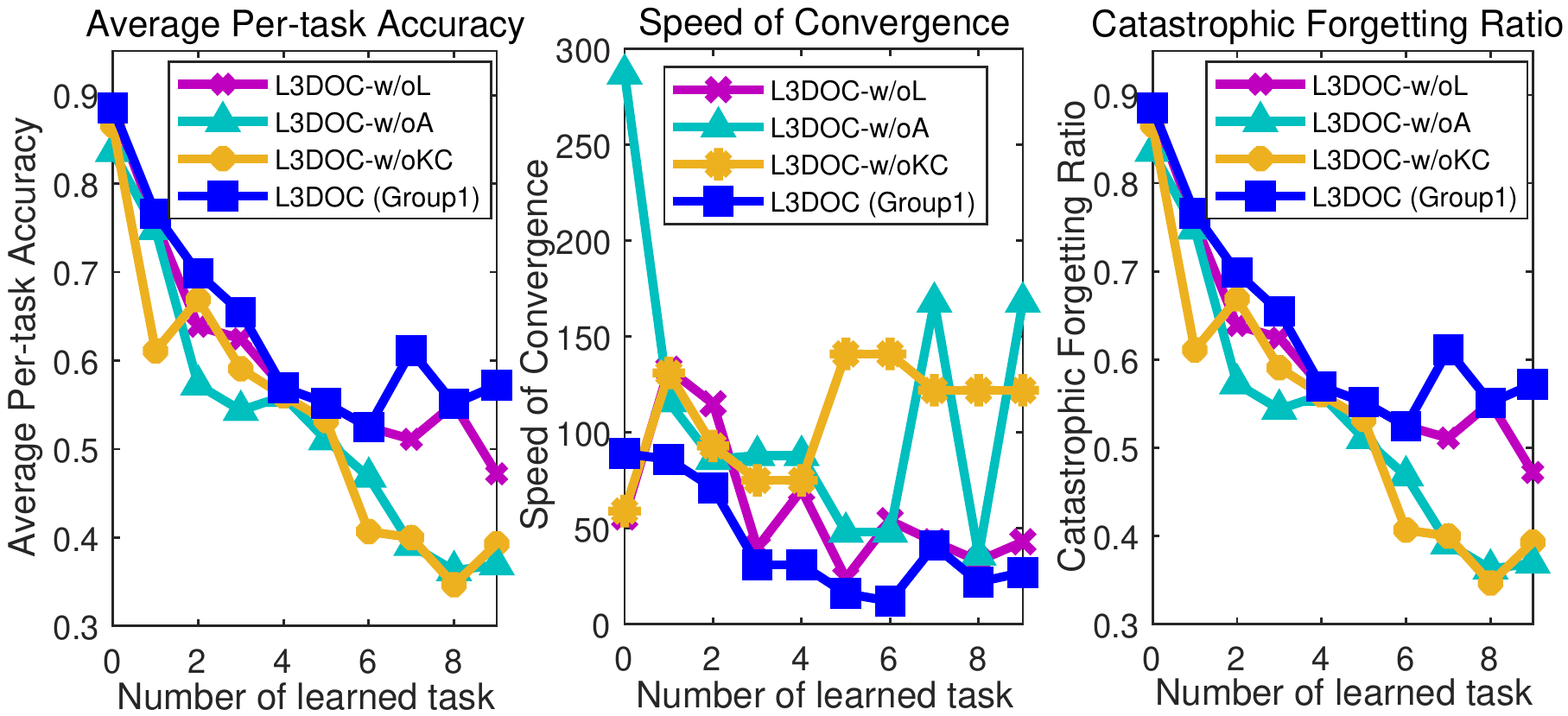}
	\end{center}
	\vspace{-15pt}
	\caption{Performance metrics on L3DOC without Memory Attention Mechanism (L3DOC-w/oA), L3DOC's Memory Attention Mechanism without knowledge gap (L3DOC-w/oL, \emph{i.e.}, $\lambda$=0 in Eq.~\eqref{eq:losstotal}), L3DOC's Memory Attention Mechanism without task-specific tensors gap (L3DOC-w/oKC, \emph{i.e.}, $\{\lambda_{K_i}\}_{i=1}^{l_{\mathrm{max}}}=0$ and $\{\lambda_{C_i}\}_{i=1}^{l_{\mathrm{max}}}=0$ in Eq.~\eqref{eq:losstotal}), and our L3DOC(Group1)) on ModelNet10. }
	\label{fig:com}
	\vspace{-8pt}
\end{figure}

\subsubsection{Effect of Memory Attention Mechanisms}
To further examine the efficiency and effectiveness of the memory attention mechanisms, as shown in Figure~\ref{fig:com}, L3DOC-w/oA, L3DOC-w/oL and L3DOC-w/oKL denote our loss function in Eq.~\eqref{eq:losstotal} without memory attention mechanisms, $\lambda$, and $\{\lambda_{K_i}\}_{i=1}^{l_{\mathrm{max}}}$ and $\{\lambda_{C_i}\}_{i=1}^{l_{\mathrm{max}}}$, respectively. Notice that the corresponding performances can degrade obviously when removing each modules, which demonstrate the effectiveness of adding memory attention mechanism (\emph{i.e.,} task-specific knowledge among previous tasks), and its rationality on the 3D classification task.

\subsubsection{Experimental results on PointNet++}
\vspace{-5pt}
To further illustrate the scalability of L3DOC, we extend our framework into PointNet++~\cite{QiYSG2017} backbone network. As the computational efficiency shown in Table~\ref{tab:expand}, our L3DOC can completes training tasks quickly and in small memory based on its layer-wise point-knowledge factorization architecture on both PointNet and PointNet++ networks. Meanwhile, for the classification accuracy, our L3DOC which is based on PointNet backbone network can achieve a more efficient performance, except for the accuracy metrics: Average-\emph{APA}. It is because the basic PointNet++ network outperforms PointNet in terms of accuracy.
\vspace{-10pt}
\begin{table}[h]
	\caption{Comparison our L3DOC with STL on the performance of average \emph{APA}, average \emph{CFR}, \emph{TT}, computational memory and the number of parameters on ModelNet10. The best performance is indicated in red.}
	\vspace{-8pt}
	\begin{center}
		\normalsize
		\setlength{\tabcolsep}{1.8mm}{
			\begin{tabular}{|l|c|c|c|c|}
				\hline
				$\#$ Backbone Network& \multicolumn{2}{|c|}{PointNet~\cite{QiSMG17}} & \multicolumn{2}{|c|}{PointNet++~\cite{QiYSG2017}}\\
				\hline
				$\#$ Framework    &  STL & L3DOC & STL & L3DOC \\
				\hline\hline
				Average-\emph{APA}  &  38.74  & 60.13  & 41.66  &  \textcolor[rgb]{1.00,0.00,0.00}{64.04} \\
				Average-\emph{CFR}  &  0.428  & \textcolor[rgb]{1.00,0.00,0.00}{0.684}  & 0.446 &  0.683    \\
				\emph{TT} (10k sec)&  1.306  & \textcolor[rgb]{1.00,0.00,0.00}{0.891}  & 34.06  &  31.57  \\
				Memory (1k Mib)       &  2.435  & \textcolor[rgb]{1.00,0.00,0.00}{1.477}  & 2.965 &  2.481  \\
				Parameter (100k)    &  15.99  & \textcolor[rgb]{1.00,0.00,0.00}{9.50}  & 79.89 &  13.25  \\
				\hline
		\end{tabular}}
	\end{center}
	\label{tab:expand}
	\vspace{-10pt}
\end{table}

%------------------------------------------------------------------------
\section{Conclusion}\label{sec:conclusion}
\vspace{-5pt}
In this paper, we propose to solve continual 3D object classification task from the perspective of lifelong learning, i.e., \underline{L}ifelong \underline{3D} \underline{O}bject \underline{C}lassification (L3DOC). To achieve lifelong learning, a layer-wise point-knowledge factorization architecture is employed in L3DOC to capture and store the shared point-knowledge, while obtaining more compact parameter representations to reduce computational complexity and memory. To efficiently avoid the catastrophic forgetting, a memory attention mechanism is used to transfer similar task-specific knowledge among different classification tasks. The experiment result on three point cloud benchmark datasets demonstrate the effectiveness and efficiency of our proposed L3DOC model when comparing with the state-of-the-arts.

{\small
\bibliographystyle{ieee_fullname}
\bibliography{3DLORbib}
}

\end{document}